\documentclass[11pt]{article}
\pdfoutput=1

\usepackage[final]{acl}

\usepackage{times}
\usepackage{latexsym}

\usepackage[T1]{fontenc}

\usepackage[utf8]{inputenc}
\usepackage{amsfonts}

\usepackage{microtype}
\usepackage{graphicx}
\usepackage{booktabs}   
\usepackage{multirow}   
\usepackage{placeins}   %
\usepackage{tcolorbox}
\usepackage{adjustbox} 
\usepackage{enumitem}  
\usepackage{amsmath}
\usepackage{listings}
\usepackage{xcolor}
\usepackage{siunitx}

\usepackage{inconsolata}

\title{RARE: Redundancy-Aware Retrieval Evaluation Framework for High-Similarity Corpora}

\author{
  Hanjun Cho \\
  Allganize\\
  \texttt{hanjun.cho@allganize.ai} \\\And
  Jay-Yoon Lee\thanks{\ Corresponding author.} \\
  Seoul National University \\
  \texttt{lee.jayyoon@snu.ac.kr}
}

\begin{document}
\renewcommand{\thefootnote}{\fnsymbol{footnote}}
\maketitle
\renewcommand{\thefootnote}{\arabic{footnote}}
\setcounter{footnote}{0}

\begin{abstract}
Existing QA benchmarks typically assume distinct documents with minimal overlap, yet real-world retrieval-augmented generation (RAG) systems operate on corpora such as financial reports, legal codes, and patents, where information is highly redundant and documents exhibit strong inter-document similarity. This mismatch undermines evaluation validity: retrievers can be unfairly undervalued even when they retrieve documents that provide sufficient evidence, because redundancy across documents is not accounted for in evaluation. On the other hand, retrievers that perform well on standard benchmarks often generalize poorly to real-world corpora with highly similar and redundant documents. We present \textbf{RARE} (Redundancy-Aware Retrieval Evaluation), a framework for constructing realistic benchmarks by (i) decomposing documents into atomic facts to enable precise redundancy tracking and (ii) enhancing LLM-based data generation with \textbf{CRRF}. RAG benchmark data usually requires multiple quality criteria, but LLMs often yield trivial outputs. CRRF scores criteria separately and fuses decisions by rank, improving the reliability of generated data. Applying RARE to Finance, Legal, and Patent corpora, we introduce \textbf{RedQA}, where a strong retriever baseline drops from 66.4\% PerfRecall@10 on 4-hop General-Wiki to 5.0--27.9\% PerfRecall@10 at 4-hop depth, revealing robustness gaps that current benchmarks fail to capture. RARE enables practitioners to build domain-specific RAG evaluations that faithfully reflect real-world deployment conditions.
\end{abstract}

\section{Introduction}

Large language models (LLMs) require external knowledge retrieval to remain accurate and current~\citep{lewis2021retrievalaugmentedgenerationknowledgeintensivenlp,izacard2022atlasfewshotlearningretrieval}. In enterprise settings, RAG operates on specialized corpora---financial disclosures, legal statutes, patent filings---that exhibit a critical property: high information redundancy and inter-document similarity.

\begin{figure}[t]
    \centering
    \includegraphics[width=\linewidth]{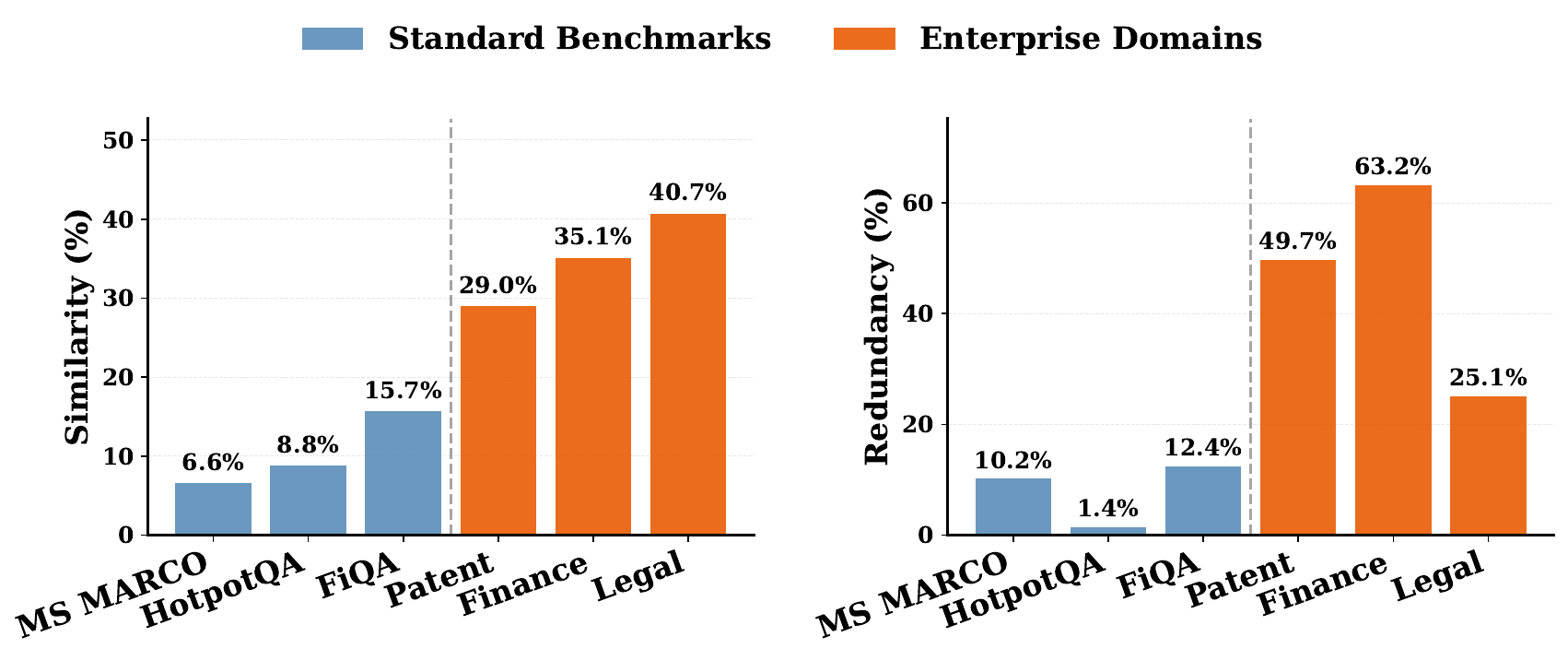}
    \caption{\textbf{Similarity (\%) and Redundancy (\%) across datasets.
    Similarity}: average pairwise cosine similarity of passage embeddings; 
    Redundancy: fraction of facts whose content also appears in a different passage.
    Standard benchmarks exhibit low values, whereas enterprise corpora show substantially higher levels.}
    \label{fig:redundancy}
\end{figure}

Standard benchmarks do not reflect specialized-domain conditions. Figure~\ref{fig:redundancy} quantifies this gap: enterprise corpora (orange) exhibit substantially greater within-corpus overlap than standard benchmarks (blue) such as Natural Questions~\citep{kwiatkowski-etal-2019-natural}, HotpotQA~\citep{yang2018hotpotqadatasetdiverseexplainable}, and FiQA~\citep{maia2018www18openchallenge}, with higher similarity and redundancy.
Here, \textbf{similarity} is the average cosine similarity between passage embeddings, and \textbf{redundancy} measures the fraction of facts whose content also appears in a different passage in the corpus~\citep{thakur2021beirheterogenousbenchmarkzeroshot}; see Appendix~\ref{app:sim-red} for metric details.

This mismatch results in two evaluation failures. First, in high-redundancy settings, retrievers that surface valid alternative evidence are penalized by single-canonical-passage labeling~\citep{zhu2025ragevalscenariospecificrag}. Second, evaluations on low-overlap benchmarks (\textit{i.e.}, those with low inter-document similarity and low fact-level redundancy) can overestimate real-world performance on high-overlap enterprise corpora, thereby obscuring robustness gaps.

Constructing benchmarks that reflect these conditions requires two capabilities. 
First, \textbf{systematic redundancy tracking}: the evaluation should recognize semantically equivalent evidence that appears in multiple passages~\citep{guha2023legalbenchcollaborativelybuiltbenchmark}. 
This is distinct from training data construction, which typically requires only local control over positive/negative pairs~\citep{karpukhin2020densepassageretrievalopendomain}; without redundancy-aware labeling, retrievers that surface valid alternatives are unfairly penalized~\citep{zhu2025ragevalscenariospecificrag}. 
Second, \textbf{high-quality data generation}: LLMs frequently produce trivial, unnatural, or contradictory questions~\citep{fu2024qgevalbenchmarkingmultidimensionalevaluation} (Appendix~\ref{app:prompts}), which undermines benchmark validity.

We introduce \textbf{RARE} (Redundancy-Aware Retrieval Evaluation), a modular framework that addresses both challenges. RARE decomposes documents into atomic facts, tracks redundancy via embedding similarity combined with LLM verification, and stabilizes data generation with \textbf{CRRF} (Criterion-wise Prompting with Reciprocal Rank Fusion), which evaluates each quality criterion independently and aggregates decisions via rank-based fusion to improve the reliability of generated benchmark data~\citep{Cormack2009ReciprocalRF}.

Experiments reveal severe degradation under realistic conditions. For 4-hop queries across the Finance, Legal, and Patent domains, strong retriever baseline attains only \textbf{27.9\%} PerfRecall@10 on Patent, with Finance and Legal dropping to \textbf{8.5\%} and \textbf{5.0\%}, respectively, whereas low-overlap General-Wiki maintains \textbf{66.4\%} at 4-hop.

\paragraph{Contributions.}
\begin{itemize}[leftmargin=*,noitemsep,topsep=3pt]
    \item We propose \textbf{RARE}, a generalizable framework that enables practitioners to construct domain-specific RAG benchmarks on their own corpora.
    \item We introduce \textbf{CRRF}, a simple multi-criteria quality-control recipe for LLM-based data generation that improves the reliability of the resulting benchmark instances.
    \item We quantify the gap between existing benchmarks and enterprise corpora via redundancy and similarity analysis, and introduce \textbf{RedQA}, a benchmark capturing these conditions across Finance, Legal, and Patent.
\end{itemize}

\section{Related Work}
\paragraph{Open-Domain QA Benchmarks.}
Existing benchmarks such as HotpotQA~\citep{yang2018hotpotqadatasetdiverseexplainable} and MS MARCO~\citep{bajaj2018msmarcohumangenerated} have played a central role in the development of open-domain question answering and retrieval models. A common design choice across these datasets is the use of clearly distinct documents with predominantly one-to-one answer--passage associations. While this setting enables clean evaluation, it reflects an idealized scenario that differs from real-world, domain-specific corpora.

\paragraph{Toward More Realistic Retrieval Benchmarks.}
Recent work has sought to increase the realism of retrieval benchmarks along multiple dimensions. MIRACL~\citep{zhang-etal-2023-miracl} extends retrieval evaluation to a multilingual setting, while other benchmarks such as KILT~\citep{petroni2021kiltbenchmarkknowledgeintensive} and TREC-COVID~\citep{voorhees2020treccovidconstructingpandemicinformation} emphasize domain-specific and knowledge-intensive retrieval scenarios. These efforts enhance realism with respect to language coverage and domain specificity; however, they do not explicitly model settings with high similarity and redundancy, where the same information is repeatedly instantiated across multiple documents in slightly varied forms.

\paragraph{General Retrieval Evaluation Suites.}
Beyond individual QA benchmarks, several evaluation suites aim to provide standardized retrieval assessment across diverse tasks and domains. BEIR~\citep{thakur2021beirheterogenousbenchmarkzeroshot} and MTEB~\citep{muennighoff2023mtebmassivetextembedding} consolidate a wide range of retrieval tasks under unified evaluation protocols, supporting systematic model comparison and reproducibility. Although these suites substantially broaden benchmark coverage, they continue to rely on fixed corpora and static relevance annotations, which limit their ability to reflect deployment-time retrieval dynamics.

\paragraph{LLM-based Data Generation.}
LLMs have recently been explored as a means of synthetic data generation. Approaches such as Promptagator~\citep{dai2022promptagatorfewshotdenseretrieval}, InPars~\citep{bonifacio2022inparsdataaugmentationinformation,jeronymo2023inparsv2largelanguagemodels}, and PAQ~\citep{lewis2021paq65millionprobablyasked} demonstrate the feasibility of automatically generating large-scale query or QA datasets. However, most existing pipelines focus on surface-level generation such as paraphrasing. Effective retrieval questions often require structured reasoning and the satisfaction of multiple quality criteria. Systematically supporting such reasoning in data generation remains an open challenge.

\begin{figure*}[t]
    \centering
    \includegraphics[width=\textwidth, height=0.55\textwidth]{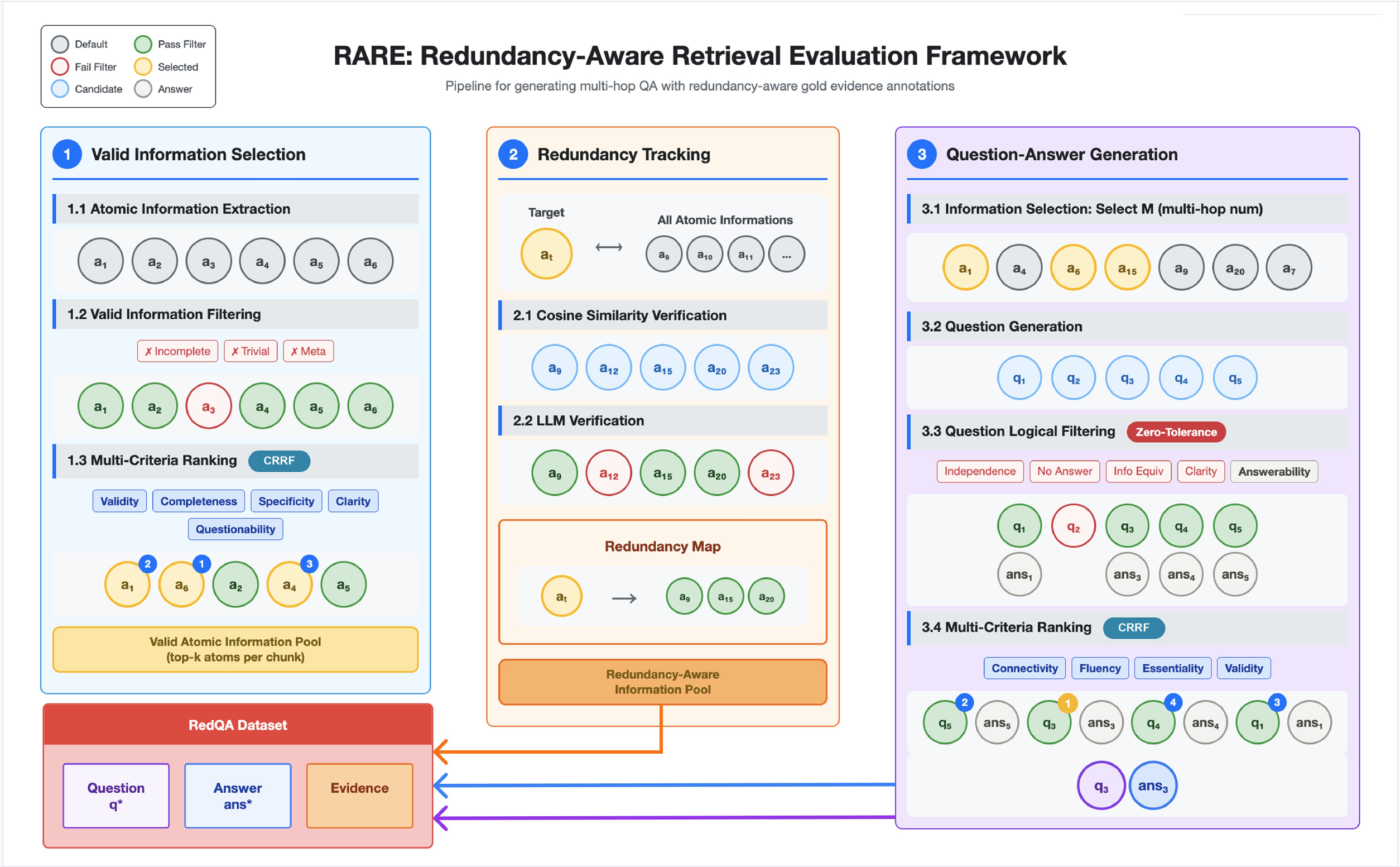}
    \caption{Overall workflow of the \textbf{RARE} framework. The pipeline consists of valid information selection, redundancy tracking, and question-answer generation. Multi-criteria judgments are stabilized through CRRF, and redundancy-aware labeling ensures fair evaluation under real-world corpus conditions. During the logical filtering step in Question-Answer Generation, answerability validation additionally performs answer generation.}
    \label{fig:framework}
\end{figure*}

\section{RARE Framework}\label{sec:method}

\subsection{Pipeline Overview}\label{sec:overview}
RARE constructs realistic multi-hop RAG evaluation data under high redundancy and similarity. Figure~\ref{fig:framework} illustrates the pipeline, which consists of three stages: \emph{Valid Information Selection}, \emph{Redundancy Tracking}, and \emph{Question--Answer Generation}. Each stage follows a common three-step pattern: constructing a candidate pool, filtering invalid candidates, and ranking candidates based on multiple criteria.

A central component of RARE is our ranking method, CRRF (detailed in Section~\ref{sec:CRRF}), which stabilizes LLM judgments by evaluating each quality criterion independently and fusing results through rank aggregation rather than confidence scores. By decomposing complex multi-criterion reasoning into simple binary or scalar judgments, CRRF yields more stable and higher-quality decisions.

\paragraph{Notation.}
\small
\begin{itemize}[noitemsep,leftmargin=12pt,topsep=0pt]
    \item $C$: input document chunk
    \item $\mathcal{A} = \{a_1, a_2, \ldots, a_n\}$: set of extracted atomic information units
    \item $\mathcal{A}_{\text{valid}} \subseteq \mathcal{A}$: atomic units passing validity filters
    \item $\mathcal{T} \subseteq \mathcal{A}_{\text{valid}}$: top-$k$ target units selected per chunk
    \item $\phi(\cdot) \in \{\textsc{Pass},\textsc{Fail}\}$: binary LLM judge
\end{itemize}
\normalsize

\begin{figure*}[t]
\centering
\includegraphics[width=\textwidth, height=0.48\textwidth]{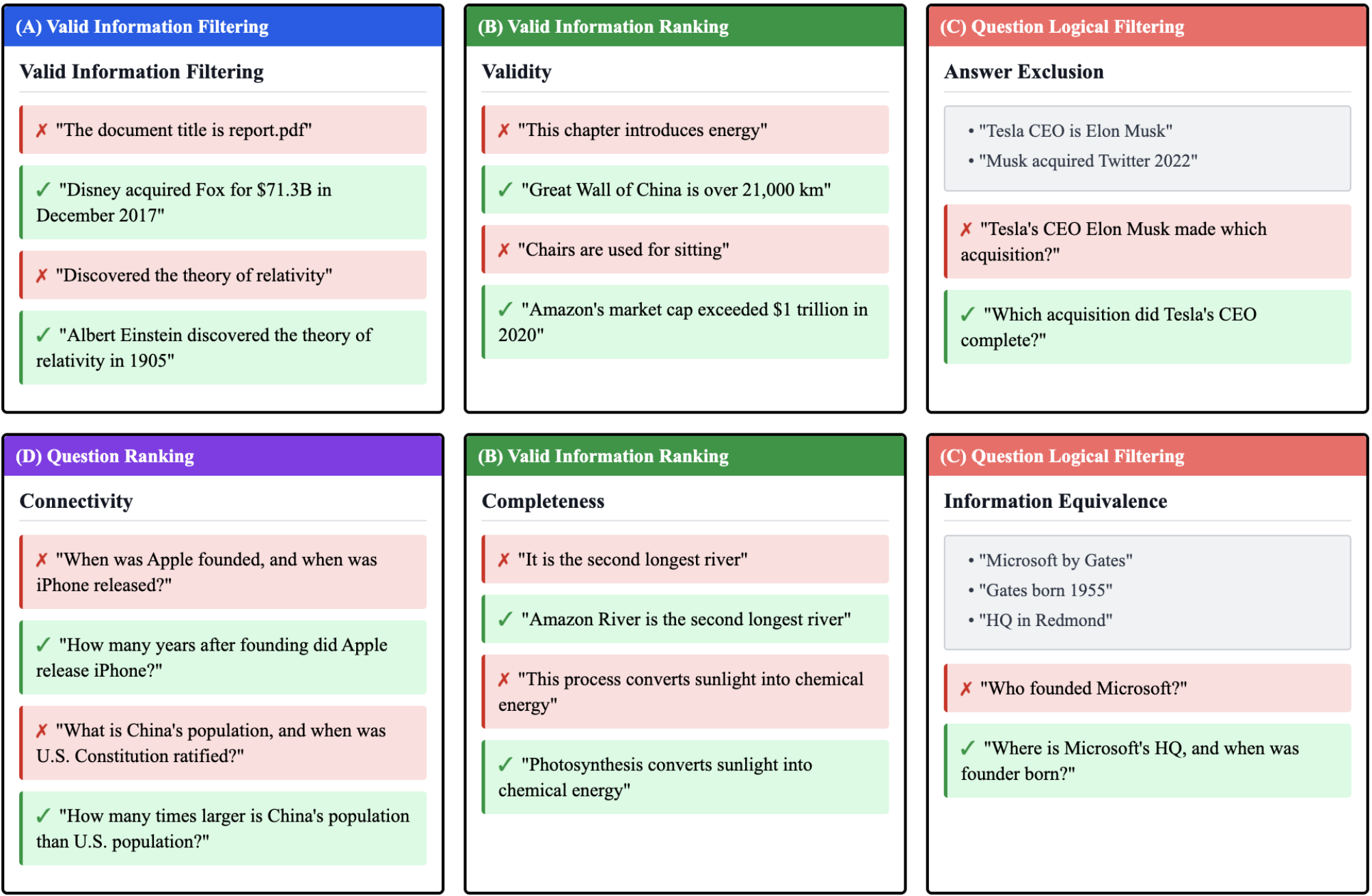}
\caption{Representative examples of filtering and ranking criteria in RARE. (A) Valid information filtering removes incomplete units lacking essential elements. (B) Two of five quality criteria for information ranking: Validity (substantive usefulness) and Completeness (self-contained clarity). (C) Two of five logical checks for questions: Answer Exclusion (no embedded answers) and Information Equivalence (exact scope matching). (D) One of four quality criteria for question ranking: Connectivity (natural logical flow). Grey boxes indicate given information required for evaluation. Complete criterion descriptions and additional examples are provided in Appendix~\ref{app:prompts}.}
\label{fig:criteria}
\end{figure*}

\subsection{Valid Information Selection}

This stage builds a reliable pool of candidate facts for multi-hop question generation through three steps: (i) decompose chunks into minimal factual claims, (ii) filter units that fail minimal validity, and (iii) prioritize the remainder by multi-criteria ranking.

\paragraph{Atomic Information Extraction.}
Each input chunk $C$ is decomposed by an LLM into atomic information units:
\[
\mathcal{A} = \{a_1,a_2,\ldots,a_n\} = f_{\text{LLM}}(C).
\]
Each $a_i$ denotes the smallest indivisible factual statement from the text. Unlike passages where multiple facts intertwine, atomic units isolate single claims. This granularity enables (i) precise cross-document redundancy tracking and (ii) modular building blocks for flexible multi-hop composition. Refer to Appendix~\ref{app:atomic-examples} and Appendix~\ref{app:prompt-atomic} for the actual examples and prompt design.

\paragraph{Validity Filtering.}
Not all extracted units are suitable for retrieval-based QA. We remove units that fail minimal validity:
\[
\mathcal{A}_{\text{valid}} = \{\, a \in \mathcal{A} \mid \phi(a) = \textsc{Pass} \,\}.
\]
Each unit is checked against three minimal criteria: (1) the unit is complete and self-contained, (2) it conveys non-trivial, utility-relevant knowledge, and (3) it expresses factual content rather than metadata or structural fragments. This step prunes incomplete or low-value units, leaving a pool of reliable factual knowledge. (Appendix~\ref{app:prompt-valid-filter})

\paragraph{Multi-Criteria Ranking.}
Passing minimal validity does not make all remaining units equally useful. We assess each atomic unit on five criteria (Appendix~\ref{app:prompt-valid-rank}; Figure~\ref{fig:criteria} for representative examples): (1) \emph{Validity}---whether the fact is substantively meaningful, (2) \emph{Completeness}---whether it is interpretable without external context, (3) \emph{Specificity}---whether it provides concrete details (names, dates, numbers), (4) \emph{Clarity}---whether it avoids ambiguity, and (5) \emph{Questionability}---whether it can naturally form a question. Criterion-wise ranks are aggregated via CRRF (Section~\ref{sec:CRRF}). From each chunk we select the top-$k$ units to construct the \emph{Valid Information Pool}, supporting redundancy tracking and subsequent question generation.

\subsection{Systematic Redundancy Tracking}

\paragraph{Problem Definition.}
In real corpora, identical factual content often appears across multiple chunks. Suppose question $q_i$ is generated from source chunk $P_i$. While the answer is contained in the source, $\mathrm{Ans}(q_i) \subseteq P_i$, it may also be supported by other chunks: $\exists j \neq i$ such that $\mathrm{Ans}(q_i) \subseteq P_j$. A retriever returning such $P_j$ is factually correct but would be penalized under naive gold annotation that assumes a single origin for each answer.

\paragraph{Two-Stage Detection.}
Starting from the \emph{Valid Information Pool}, we select the top-$k$ atomic units from each chunk to form the target set $\mathcal{T} \subseteq \mathcal{A}_{\text{valid}}$. For every target $a_t \in \mathcal{T}$, we first retrieve a high-recall candidate set using embedding similarity:
\[
\begin{aligned}
\mathcal{C}_\tau(a_t)
= \bigl\{\, a_j \in \mathcal{A} \ \bigm|\ 
&\mathrm{chunk}(a_j)\neq \mathrm{chunk}(a_t),\\
&\text{sim}(a_t,a_j) \ge \tau \,\bigr\}.
\end{aligned}
\]
where $\text{chunk}(a_t)$ denotes the originating chunk. We then refine candidates with an LLM judge $\phi(a_t,a_j)\in\{\textsc{Pass},\textsc{Fail}\}$, where \textsc{Pass} indicates that $a_t$ and $a_j$ are factually equivalent (i.e., redundant), and \textsc{Fail} indicates non-equivalence:
\[
\mathcal{R}(a_t) = \{\, a_j \in \mathcal{C}_\tau(a_t) \mid \phi(a_t,a_j) = \textsc{Pass} \,\}.
\]
Atomic granularity reduces noise in the embedding space and narrows the gap between semantic similarity and factual equivalence, making equivalence checks more tractable for LLMs. Embedding similarity search uses a recall-focused threshold ($\tau=0.5$) to emphasize recall, while LLM verification enforces precision. Redundancy labels are recorded as key--value mappings $(a_t \mapsto \mathcal{R}(a_t))$ for all targets.

\subsection{Question-Answer Generation}

Candidate facts are combined into reasoning chains, filtered for logical soundness, and ranked by multi-criterion judgments to yield challenging yet well-formed multi-hop items.

\paragraph{Question Candidate Construction.}
From the redundancy-aware valid pool, we randomly sample $N$ atomic units to form input set $A_N \subset \mathcal{A}_{\text{valid}}$. Within this set, the LLM selects exactly $M$ units with strong relational connectivity, encouraging connected reasoning rather than mere fact listing:
\[
T = f_{\text{select}}(A_N, M), \quad T \subseteq A_N,\; |T|=M.
\]
Using these $M$ units as common evidence, we generate $S$ paraphrased candidate questions that all require the same underlying evidence:
\[
Q = \{\, q_1, q_2, \ldots, q_S \,\} = f_{\text{gen}}(T,S).
\]
Here, $N$ denotes the number of sampled atoms, $M$ the number of hops (i.e., the number of distinct atomic units that must be jointly retrieved to answer the question), and $S$ the number of candidate questions. Producing multiple candidates mitigates generation-time errors by enabling later filtering and ranking to discard flawed questions while retaining high-quality ones.

\paragraph{Logical Filtering (Zero-Tolerance).}
Because LLMs may introduce diverse logical errors during generation, we apply strict filtering based on five criteria (Appendix~\ref{app:prompt-qfilter}): (1) \emph{Contextual Independence}---the question must not rely on document structure or meta-references, (2) \emph{Answer Exclusion}---no answer content or intermediate reasoning may be embedded in the question, (3) \emph{Information Equivalence}---the informational scope of the question must \emph{exactly} match the designated atomic evidence set (requiring all and only the given units), (4) \emph{Question Clarity}---the formulation must avoid ambiguity or underspecified references, and (5) \emph{Answerability}---the selected evidence must be substantively sufficient to yield a clear and unique answer.

This is enforced under a \textbf{zero-tolerance policy}: if a candidate fails even one criterion, it is immediately discarded. Formally, given $Q=\{q_1,\ldots,q_S\}$,
\[
Q^{\ast} = \{\, q \in Q \mid \forall \ell \in \mathcal{L},\; \phi_\ell(q) = \textsc{Pass} \,\},
\]
where $\mathcal{L}$ is the set of five criteria and $\phi_\ell$ is the LLM Judge for criterion $\ell$.

\paragraph{Multi-Criteria Ranking.}
To select the best question, each $q \in Q^{\ast}$ is evaluated on four criteria (Appendix~\ref{app:prompt-qrank}): (1) \emph{Connectivity}---whether the selected facts form a coherent reasoning chain, (2) \emph{Fluency}---whether the question is expressed in natural language, (3) \emph{Essentiality}---whether it focuses on core information without redundancy, and (4) \emph{Validity}---whether it asks for substantively meaningful content. Criterion-wise ranks are aggregated via CRRF (Section~\ref{sec:CRRF}), and the highest-ranked question is retained as the benchmark item.

\subsection{CRRF: Stabilizing Multi-Criteria Ranking}\label{sec:CRRF}

\paragraph{Motivation.}
Ranking candidates by multiple quality criteria is challenging for LLMs. Jointly reasoning over criteria produces unstable outputs because it requires the model to balance competing objectives simultaneously. Moreover, LLM confidence scores are unreliable and poorly calibrated across criteria, making score-based aggregation prone to magnitude mismatches.

\paragraph{Method.}
CRRF addresses both issues through two principles: (1) evaluate each criterion independently via separate prompts, and (2) fuse results using rank-based aggregation, discarding confidence magnitudes entirely.

Formally, given candidate $x$ and $N$ criteria, we obtain per-criterion ranks $\mathrm{rank}_i(x)$ through independent LLM calls. The fused score is:
{\setlength{\abovedisplayskip}{4pt}
 \setlength{\belowdisplayskip}{4pt}
\[
s(x) = \sum_{i=1}^{N} \frac{1}{\mathrm{rank}_i(x)}
\]
}
Higher $s(x)$ indicates broader inter-criterion consensus. Unlike confidence-weighted averaging, this approach relies only on ordinal preferences, which LLMs produce more reliably than calibrated probabilities.

\paragraph{Application in RARE.}
CRRF is applied at two stages: (1) ranking atomic information units by 5 criteria (validity, completeness, specificity, clarity, questionability), and (2) ranking candidate questions by 4 criteria (connectivity, fluency, essentiality, validity). Both Section~\ref{sec:ablation} and Appendix~\ref{app:CRRF} indicate that CRRF consistently achieves higher quality and stability.

\subsection{Benchmark Construction: RedQA}
The final outcome of RARE is \textbf{RedQA}, a redundancy-aware QA benchmark. Each instance consists of a top-ranked question, its reference answer, and a redundancy-aware gold evidence set. The gold set includes both the originating passages and all semantically equivalent alternatives identified through redundancy tracking, thereby reducing penalties when retrievers return correct but non-origin evidence.

\section{Dataset}\label{sec:dataset}
We evaluate RARE on four domains. Finance, Legal, and Patent represent high-overlap corpora, while General-Wiki serves as a low-overlap baseline. Table~\ref{tab:dataset-stats} summarizes corpus statistics.

\paragraph{Corpora.}
We select three specialized domains---\textbf{Finance} (SEC 10-K, 2023--2024), \textbf{Legal} (2023 U.S.\ Code), and \textbf{Patent} (USPTO, 2023--2024)---because they are widely used in enterprise RAG, exhibit naturally occurring document overlap, and are publicly accessible with clear provenance. \textbf{General-Wiki} adopts the Wikipedia corpus from
HotpotQA~\citep{yang2018hotpotqadatasetdiverseexplainable} as its document source, providing a canonical general-domain, low-overlap setting, consistent with standard Wikipedia-based
benchmarks such as Natural Questions~\citep{kwiatkowski-etal-2019-natural}, KILT~\citep{petroni2021kiltbenchmarkknowledgeintensive},
and MIRACL~\citep{zhang-etal-2023-miracl}. We use it as a representative
low-overlap baseline to isolate the effect of corpus redundancy on retrieval
robustness. See Appendix~\ref{app:domains} for details.

\paragraph{Construction.}
We use GPT-5 Nano for LLM judgments and GPT-5 for question generation; text-embedding-3-large is used for similarity computations. Multi-hop questions span 1--4 hops. Full implementation details are in Appendix~\ref{app:impl}.

\begin{table}[h]
\centering
\caption{\textbf{RedQA dataset statistics}. Chunks denote retrieval units in the reference corpus, and Samples denote QA instances. Enterprise corpora exhibit substantially higher redundancy and similarity than General-Wiki.}
\label{tab:dataset-stats}
\begin{adjustbox}{max width=\linewidth}
\small
\begin{tabular}{lrrrr}
\toprule
\textbf{Corpus} & \textbf{Chunks} & \textbf{Redundancy (\%)} & \textbf{Similarity (\%)} & \textbf{Samples} \\
\midrule
Finance  & 3{,}281 & 63.2 & 35.1 & 645 \\
Patent   & 3{,}339 & 49.7 & 29.0 & 480 \\
Legal    & 4{,}090 & 25.1 & 40.7 & 605 \\
General-Wiki & 3{,}218 & 1.4  & 8.8  & 688 \\
\bottomrule
\end{tabular}
\end{adjustbox}
\end{table}

\subsection{Human Evaluation of LLM Filtering}
\label{sec:human-eval}
We validated two filtering stages---\emph{Information Filtering} and \emph{Question Logical Filtering}---on 480 items, evenly sampled across hop depths (1--4), domains (Finance/Patent/Legal), and pass/fail strata. Eight NLP researchers independently judged 60 items each using the same rubric as the LLM; about 17 items were adjudicated via group discussion. All assessments were conducted as part of their regular duties. Human judgments served as gold labels. Throughout, we treat \emph{Fail} as the positive class.

\begin{table}[t]
\centering
\caption{\textbf{Human evaluation statistics}. \emph{Fail} is treated as the positive class. The approach prioritizes recall over precision to minimize false negatives, yielding only 4.2\% false negatives overall.}
\label{tab:human-eval}
\begin{adjustbox}{max width=\columnwidth}
\begin{tabular}{lccccccc}
\toprule
\textbf{Stage} & \textbf{TP} & \textbf{FP} & \textbf{FN} & \textbf{TN} & \textbf{Prec.} & \textbf{Rec.} & \textbf{Acc.} \\
\midrule
Information Filtering & 69 & 51 & 8 & 112 & 57.5\% & \textbf{89.6\%} & 75.4\% \\
Question Filtering & 61 & 59 & 12 & 108 & 50.8\% & \textbf{83.6\%} & 70.4\% \\
\bottomrule
\end{tabular}
\end{adjustbox}
\end{table}

Our filtering strategy achieves high recall (89.6\% and 83.6\%) at the cost of moderate precision (57.5\% and 50.8\%), resulting in overall accuracy of 75.4\% and 70.4\%. Only \textbf{20 false negatives (4.2\%)} occurred across all 480 samples. This design reflects a key principle for benchmark construction: excluding some valid content is preferable to retaining logically flawed questions, as the latter would directly compromise evaluation validity.

\section{Experiments}\label{sec:exp}
\subsection{Setup}
\paragraph{Models.}
We compare a sparse retriever, BM25~\citep{robertson2009bm25}, against dense retrievers spanning both native embedding models and LLM-based variants. The commercial baseline is OpenAI-Large (text-embedding-3-large). Open-source models include pure embedding architectures (E5-Large~\citep{wang2024multilinguale5textembeddings}, BGE-M3~\citep{chen2024bgem3embeddingmultilingualmultifunctionality}, Jina-v4~\citep{gunther2025jinaembeddingsv4universalembeddingsmultimodal}) and LLM-derived embeddings (E5-Mistral-7B~\citep{wang-etal-2024-improving-text,jiang2023mistral7b}, Qwen3-Embedding variants~\citep{zhang2025qwen3embeddingadvancingtext,yang2025qwen3technicalreport}). This suite enables comparisons across sparse vs.\ dense paradigms, parameter scales, and embedding model families.

\begin{table*}[t]
\centering
\resizebox{\textwidth}{!}{%
\begin{tabular}{l|c|cc|cc|cc|cc}
\toprule
&  & \multicolumn{2}{c|}{\textbf{Finance}} 
& \multicolumn{2}{c|}{\textbf{Patent}} 
& \multicolumn{2}{c|}{\textbf{Legal}} 
& \multicolumn{2}{c}{\textbf{General-Wiki}} \\
\cmidrule(lr){3-4}\cmidrule(lr){5-6}\cmidrule(lr){7-8}\cmidrule(lr){9-10}
\textbf{Model} & \textbf{Size} 
& Coverage@10 & PerfRecall@10
& Coverage@10 & PerfRecall@10
& Coverage@10 & PerfRecall@10
& Coverage@10 & PerfRecall@10 \\
\midrule
BM25            & -- & 59.87 & 36.12 & 76.35 & 55.63 & 49.08 & 27.77 & 64.58 & 44.19 \\
\midrule
OpenAI-Large    & -- & 67.67 & 42.95 & 81.96 & 61.46 & 61.17 & 37.36 & 92.78 & 86.48 \\
\midrule
E5-Large        & 0.56B & 62.83 & 37.98 & 80.21 & 60.62 & 52.56 & 30.91 & 91.67 & 84.01 \\
BGE-M3          & 0.56B & 64.43 & 39.84 & 81.82 & 59.58 & 54.50 & 32.40 & 90.60 & 82.27 \\
Qwen3           & 0.6B  & 67.62 & 43.88 & 81.49 & 60.21 & 60.29 & 35.37 & 89.96 & 78.78 \\
\midrule
Jina-v4         & 3.75B & 65.10 & 40.31 & 80.09 & 59.58 & 59.57 & 34.55 & 92.01 & 85.03 \\
Qwen3           & 4B    & 72.43 & \textbf{48.84} & 83.09 & 62.50 & 65.12 & 39.67 & \textbf{93.97} & \textbf{89.83} \\
\midrule
E5-Mistral      & 7B    & 69.69 & 44.81 & 81.67 & 62.71 & 56.86 & 34.05 & 93.53 & 88.66 \\
Qwen3           & 8B    & \textbf{72.92} & 47.44 & \textbf{84.05} & \textbf{63.12} & \textbf{67.16} & \textbf{41.49} & 93.58 & 88.66 \\
\bottomrule
\end{tabular}}
\caption{\textbf{Retrieval performance across domains}. Dense retrievers outperform BM25, but their advantage narrows in higher-overlap corpora (Finance, Legal, Patent). Scaling generally improves performance; however, PerfRecall@10 remains low in Finance and Legal, indicating that high-overlap settings are substantially more challenging.}
\label{tab:main-results}
\end{table*}

\paragraph{Evaluation Metrics.}
We use \textbf{Coverage@10} and \textbf{PerfRecall@10} as primary metrics (NDCG@10 and MRR in Appendix~\ref{app:results}); required information is defined at the \emph{chunk} level, and all metrics use redundancy-aware gold labels.
\begin{itemize}[leftmargin=*,noitemsep,topsep=3pt]
    \item \textbf{Coverage@K}: Fraction of required information retrieved in the Top-K (e.g., 2/4 = 50\%), measuring partial success.
    \item \textbf{PerfRecall@K}: Binary success if all required information appears in the Top-K (e.g., 2/4 = fail), indicating whether the retrieved context is sufficient to answer the question.
\end{itemize}

\subsection{Main Results}
\label{exp:main-results}

Table~\ref{tab:main-results} presents retrieval performance across four domains with varying redundancy and similarity characteristics.

\paragraph{Sparse vs.\ Dense Retrieval.}
Dense retrievers outperform BM25 across all domains, but the gap narrows in high-overlap domains. For example, in General-Wiki the Coverage@10 gap is 28.2 percentage points (BM25: 64.58\% vs.\ OpenAI-Large: 92.78\%), whereas it is 7.8 pp in Finance and 12.1 pp in Legal. This pattern is consistent with the hypothesis that dense retrieval offers less advantage when candidate passages exhibit stronger overlap.

\paragraph{Scaling Effects.}
Retrieval quality generally improves with model size. Within Qwen3, PerfRecall@10 increases with scale across domains (e.g., Legal: 35.37\% at 0.6B to 41.49\% at 8B), with similar trends in Finance and Patent (+3.56 pp and +2.9 pp, respectively). Nevertheless, even the largest models reach only 41.49\% PerfRecall@10 on Legal, indicating that high-overlap domains remain challenging under our benchmark settings.

\paragraph{LLM-based vs.\ Embedding-only.}
In our suite, LLM-based embedding models tend to outperform embedding-only baselines at comparable parameter scales. On Legal at $\approx$0.6B parameters, Qwen3-0.6B achieves higher Coverage@10 than BGE-M3 by +5.79 pp; at $\approx$4B, Qwen3-4B similarly exceeds Jina-v4 by +5.6 pp. One possible explanation is that LLM-based embeddings better capture relational and compositional semantics, which can be beneficial for multi-hop retrieval where evidence must form coherent chains. Nonetheless, the observed gaps may be confounded by differences in training and overall model capability.

\begin{figure}[h]
    \centering
    \includegraphics[width=\linewidth]{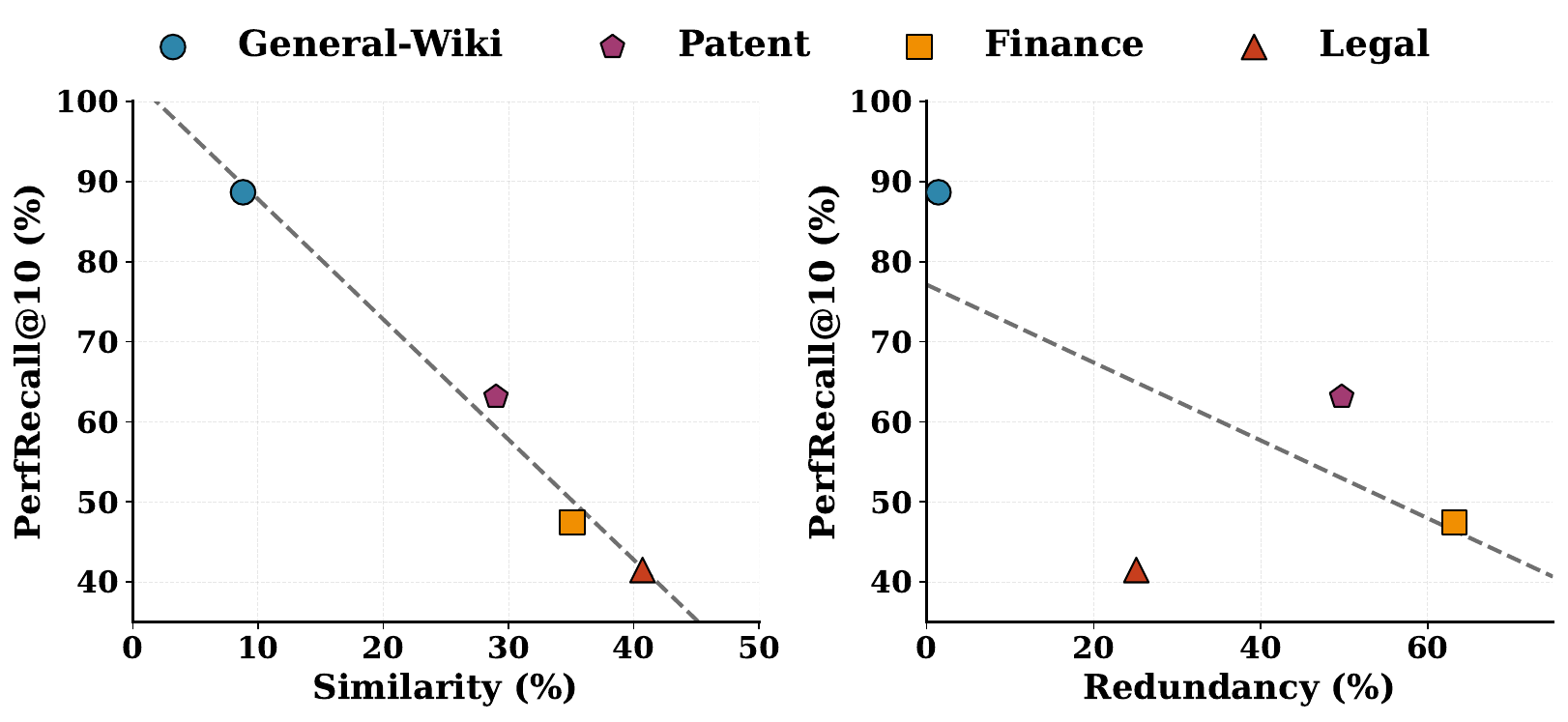}
    \caption{\textbf{PerfRecall@10 vs.\ document similarity (left) and redundancy (right) for Qwen3-8B}. PerfRecall@10 decreases monotonically with similarity, whereas redundancy shows a weaker, non-monotonic association.}
    \label{fig:correlation}
\end{figure}

\paragraph{Similarity vs.\ Redundancy.}
We examine which corpus factor better explains PerfRecall@10 degradation by contrasting domains with different redundancy and similarity profiles (Figure~\ref{fig:correlation}). While all three enterprise domains exhibit substantially higher overlap than General-Wiki, we leverage their relative differences in similarity and redundancy to disentangle which factor better predicts PerfRecall@10 degradation. Legal combines the highest similarity (40.7\%) with the lowest redundancy (25.1\%), Finance is most redundant (63.2\%) with moderate similarity (35.1\%), and Patent has the lowest similarity (29.0\%) with intermediate redundancy (49.7\%). With Qwen3-8B, PerfRecall@10 tracks similarity rather than redundancy: Legal is lowest (41.49\%), Finance intermediate (47.44\%), and Patent highest (63.12\%). This suggests that similarity drives retrieval difficulty by increasing confusion among near-duplicates, whereas redundancy can mitigate errors by providing alternative evidence paths.

\subsection{Performance Degradation with Hop Depth}
\begin{figure}[h]
    \centering
    \includegraphics[width=\linewidth]{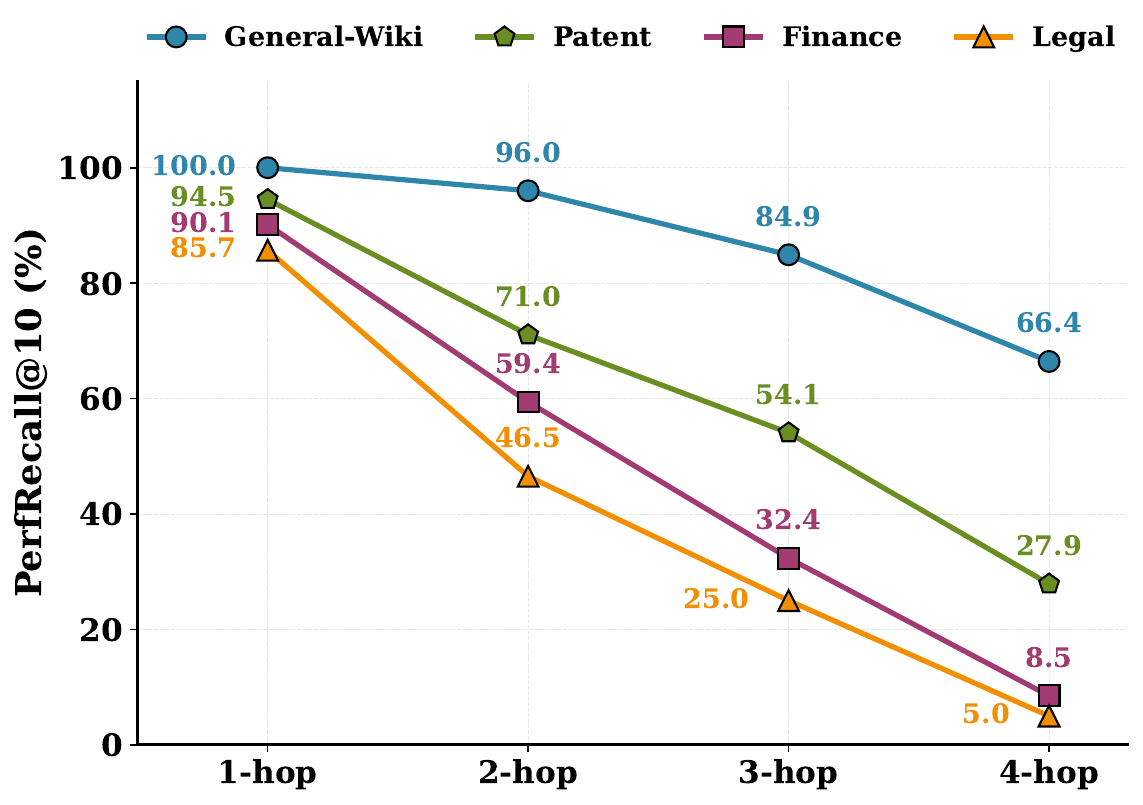}
    \caption{\textbf{Hop-wise PerfRecall@10 degradation (Qwen3-8B).} Performance degrades with increasing hop depth across all domains, with substantially sharper drops in high-overlap corpora (Finance, Legal, Patent) than in General-Wiki.}
    \label{fig:hop-analysis-scaling}
\end{figure}

Figure~\ref{fig:hop-analysis-scaling} shows how dense retrieval degrades as hop depth increases from 1 to 4 (Appendix~\ref{app:results}). On General-Wiki, Qwen3-8B shows a notable PerfRecall@10 drop from 1$\rightarrow$4 hops (\(33.6\) pp) yet remains substantially above 4-hop accuracy (66.4\%).
In contrast, high-overlap domains collapse with depth: in Finance, PerfRecall@10 drops from 90.1\% to just 8.5\% (1$\rightarrow$4 hops), an 81.6 pp drop.
We attribute this to an \textit{overlap-induced evidence bottleneck}---near-duplicate corpora cause top-$k$ retrieval to recycle facts instead of adding complementary evidence, limiting coverage and leaving the compositional evidence needed for the answer missing.
Overall, overlap amplifies multi-hop difficulty, motivating retrieval that explicitly promotes diversity across hops.

\subsection{Effectiveness of CRRF}\label{sec:ablation}
We ablate whether decomposing multi-criterion judgments and RRF improves data quality over joint prompting and score aggregation.
We evaluate (i) prompting strategy---Vanilla (holistic), Combined (joint criteria), and Separate (one criterion per prompt)---and
(ii) aggregation---Base (mean), MinMax (normalized mean), and RRF (rank fusion).
Using 100 long HotpotQA~\citep{yang2018hotpotqadatasetdiverseexplainable} passages to increase ranking difficulty, we rank sentences by the five criteria used in Valid Information Selection (validity, completeness, specificity, clarity, questionability).
Quality is measured by how highly human-annotated evidence sentences are ranked. We report NDCG@3 (MRR/Top-1 in Appendix~\ref{app:CRRF}).

Figure~\ref{fig:CRRF-perf} shows that CRRF (Separate+RRF) achieves the best NDCG@3 for both GPT-5 Nano (0.463) and GPT-5 (0.467).
Three patterns explain the gains: (i) making criteria explicit helps (e.g., Combined+RRF improves over Vanilla: 0.419 vs.\ 0.352),
(ii) separating criteria further improves over joint evaluation (0.463 vs.\ 0.419), consistent with reduced cross-criterion interference, and
(iii) rank fusion outperforms score aggregation under separate prompting (0.463 vs.\ 0.391), suggesting robustness to poorly calibrated confidence magnitudes.
Overall, CRRF provides a simple and reliable recipe for stabilizing multi-criterion LLM judgments in RARE framework.

\begin{figure}[h]
    \centering
    \includegraphics[width=\columnwidth]{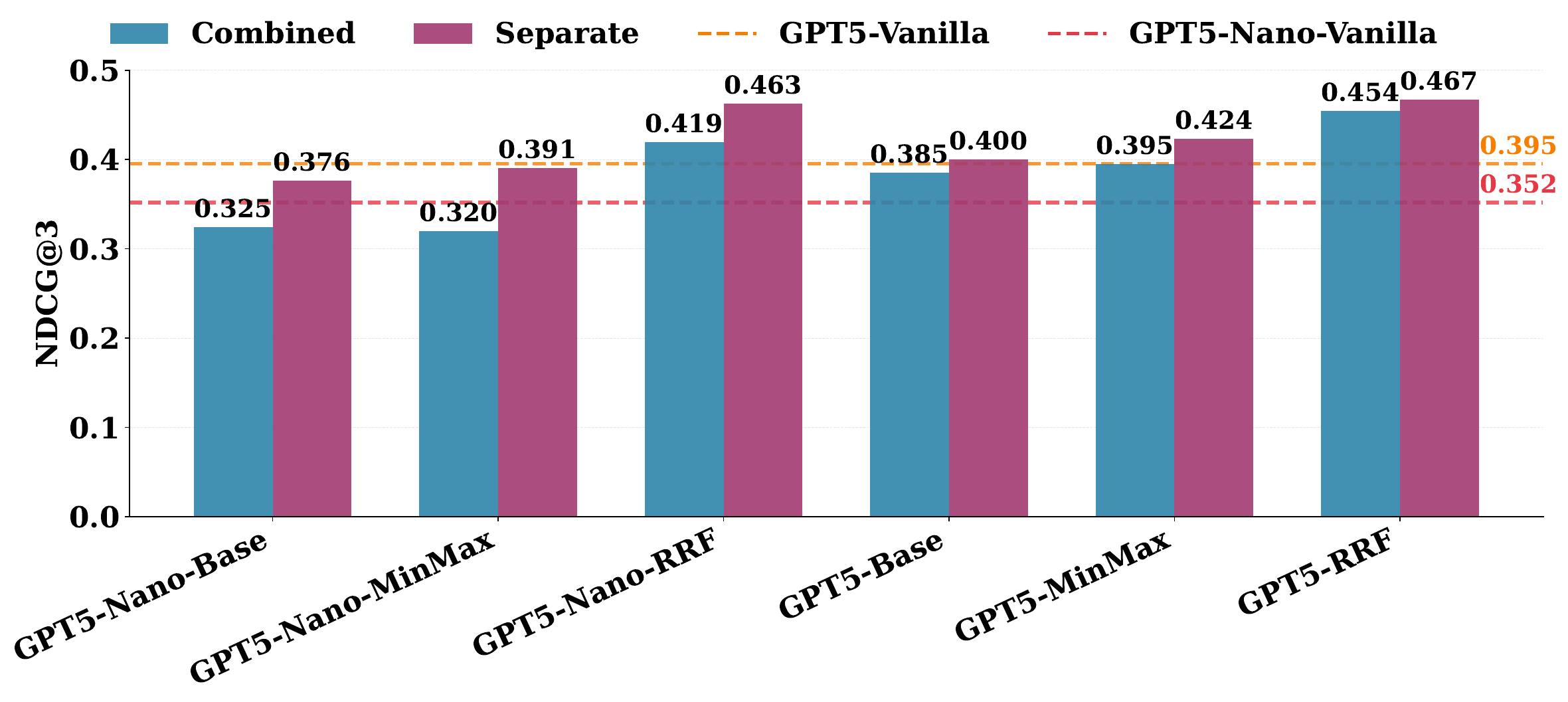}
    \caption{\textbf{CRRF performance across prompting and aggregation strategies.}
    NDCG@3 for GPT-5 Nano and GPT-5 under different prompting strategies and aggregation methods.
    CRRF (Separate + RRF) achieves the best performance, and RRF consistently outperforms score-based aggregation in this setting.}
    \label{fig:CRRF-perf}
\end{figure}

\subsection{End-to-End RAG Evaluation}\label{sec:e2e}

Prior sections evaluated retrieval in isolation, but end-to-end RAG accuracy has three determinants: retrieval quality, the generator's ability to utilize retrieved context, and its parametric knowledge. We analyze how these combine per domain on RedQA, using GPT-5 Mini as generator and a GPT-5 judge scoring answers against ground truth.

\begin{figure*}[t]
\centering
\includegraphics[width=\textwidth, height=0.35\textwidth]{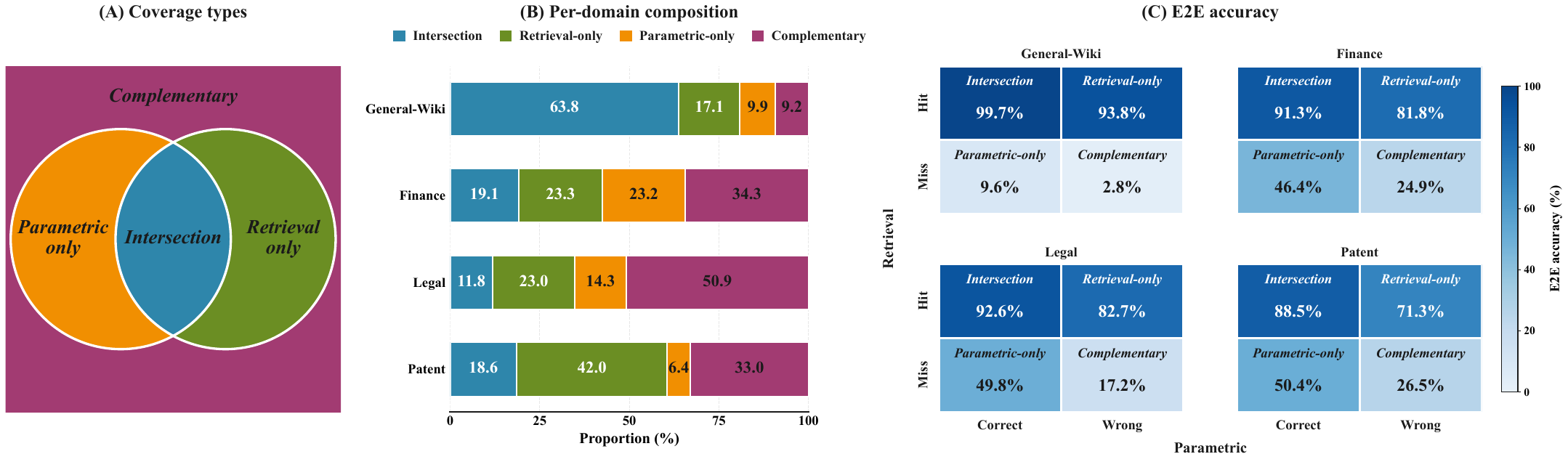}
\caption{\textbf{Coverage-type decomposition of end-to-end accuracy.} Four coverage types defined by Parametric-only correctness $\times$ perfect retrieval coverage (PerfRecall@10$=$1). (A) Schematic. (B) Per-domain share. (C) Accuracy per type: \textit{Retrieval-only} isolates utilization, \textit{Parametric-only} retention, and \textit{Complementary} fusion.}
\label{fig:e2e}
\end{figure*}

\begin{table}[h]
\centering
\small
\setlength{\tabcolsep}{4pt}
\resizebox{\columnwidth}{!}{%
\begin{tabular}{l | c | cc | cc}
\toprule
 & & \multicolumn{2}{|c}{\textbf{Parametric}} & \multicolumn{2}{|c}{\textbf{Retrieval}} \\
\cmidrule(lr){3-4}\cmidrule(lr){5-6}
\textbf{Domain} & \textbf{E2E} & \textbf{Param-only} & \textbf{RAG Gain} & \textbf{PerfRecall@10} & \textbf{Parametric Gain} \\
\midrule
General-Wiki & 80.81 & 73.69 & +7.12  & 80.88 & -0.07  \\
Finance      & 55.85 & 42.33 & +13.52 & 42.46 & +13.39 \\
Legal        & 45.82 & 26.12 & +19.70 & 34.84 & +10.98 \\
Patent       & 58.40 & 25.00 & +33.40 & 60.60 & -2.20  \\
\bottomrule
\end{tabular}}
\caption{\textbf{End-to-end (E2E) RAG accuracy across domains (\%).} Averaged over the nine retrievers in Table~\ref{tab:main-results}. Param-only: GPT-5 Mini generation without retrieval. \textit{RAG Gain} (E2E $-$ Param-only): lift retrieval adds to the parametric baseline. \textit{Parametric Gain} (E2E $-$ PerfRecall@10): lift parametric knowledge adds beyond retrieval coverage.}
\label{tab:e2e-main}
\end{table}

\paragraph{RAG Gain \& Parametric Gain.}
RAG Gain is substantial in every domain (+7.12~pp to +33.40~pp) and grows largest where parametric knowledge is weakest. Parametric Gain is strongly positive on Finance (+13.39~pp) and Legal (+10.98~pp), near zero on General-Wiki, and slightly negative on Patent.

\paragraph{Context utilization.}
\textit{Retrieval-only} accuracy isolates utilization. It is highest on General-Wiki, comparable on Finance and Legal (82--83\%), and lowest on Patent (71.3\%): the generator struggles to exploit technical Patent evidence even when retrieval hits.

\paragraph{Parametric retention.}
\textit{Parametric-only} accuracy isolates retention under retrieval miss. Over-abstention appears across all domains but is moderate on enterprise domains (around 50\%) and extreme on General-Wiki (9.6\%): when retrieved context is present, even imperfectly, the generator defers to it rather than to its own priors.

\paragraph{Signal fusion.}
\textit{Complementary} accuracy captures queries where neither source alone is sufficient to generate the answer. It remains non-trivial (e.g., Finance 24.9\%), showing that the incomplete signals complement each other to yield correct answers.

\paragraph{Gain attribution.}
These profiles explain the Gain signs in Table~\ref{tab:e2e-main}. On General-Wiki, parametric knowledge and perfectly retrieved evidence overlap heavily, limiting both Gains (RAG Gain +7.12~pp, Parametric Gain $-$0.07~pp). On Patent, the utilization bottleneck outweighs the parametric-recovery paths (\textit{Parametric-only}, \textit{Complementary}), producing a negative Parametric Gain ($-$2.20~pp).

\paragraph{Retrieval dominance.}
Across all domains, hit-cell accuracy (\textit{Intersection}, \textit{Retrieval-only}) far exceeds miss-cell accuracy, confirming retrieval quality as the dominant lever in end-to-end RAG.

\section{Conclusion}
Our work revisits an implicit assumption of retrieval evaluation: many benchmarks are built on corpora where relevant evidence is relatively distinct, whereas real-world corpora (e.g., Finance, Legal, and Patent) frequently exhibit both high redundancy and similarity. This mismatch can change retrieval difficulty and can mislead when benchmark performance is used as a proxy for deployment robustness. We propose \textbf{RARE}, a modular framework for constructing retrieval benchmarks that explicitly quantify and incorporate redundancy and passage similarity. RARE decomposes documents into atomic facts to support redundancy tracking, and employs \textbf{CRRF}, a criterion-separated rank-fusion strategy, to make LLM-based quality judgments more stable. Applying RARE to domain-specific corpora yields \textbf{RedQA}, an evaluation suite intended to better approximate enterprise-like retrieval conditions. Across domains and retriever families, we observe that retrieval quality can degrade markedly in higher-overlap settings compared to existing benchmarks. In our analyses, higher document similarity is associated with larger performance drops, while redundancy can sometimes provide additional retrieval pathways. Together, these results suggest that existing benchmarks may overestimate robustness under real-world corpus conditions, motivating evaluation protocols that consider corpus overlap.

\section*{Limitations}
RARE relies on LLM judges for generation and verification, inheriting model-specific biases and calibration issues. Our redundancy detection uses embedding similarity with a fixed threshold (\(\tau=0.5\)); while effective in our experiments, optimal settings may vary by domain. CRRF is validated on sentence ranking in Wikipedia passages; generalization to other granularities or tasks remains to be established. As hop depth increases, some generated questions become list-like despite being logically valid; alternative formulations (e.g., graph-based connectivity modeling) may help produce more natural multi-hop chains. Finally, our model comparisons do not fully control for training data across embedding families, limiting causal attribution of performance differences to architecture alone.

\section*{Ethics Statement}
This work introduces a benchmark for evaluating retrieval systems in high-redundancy environments. The dataset construction uses publicly available documents (SEC filings, USPTO patents, U.S. Code, Wikipedia) and does not involve sensitive data. LLM-based generation and filtering may inherit biases from the underlying models (GPT-5, GPT-5 Nano); we mitigate this through human evaluation (Section~\ref{sec:human-eval}) showing 89.6\%, 83.6\% recall in detecting invalid content. RedQA is intended for research purposes to improve information retrieval systems, with potential societal benefits in enterprise search, legal research, and knowledge management. All data sources are used in accordance with their intended purposes and licensing terms. AI tools were used for grammar correction, proofreading, and English composition. However, all intellectual contributions, experimental design, and analysis were entirely conducted by humans.

\section*{Acknowledgments}
This work was supported in part by the National Research Foundation of Korea (NRF) grant (RS-2023-00280883, RS-2023-00222663); by the National Research Foundation, Korea, under project BK21 FOUR (Dept.\ of Data Science, SNU, No.\ 5199990914569); by the Korea Institute of Science and Technology Information (KISTI) in 2026 (No.\ (KISTI)K26L3M1C1), aimed at developing KONI (KISTI Open Neural Intelligence), a large language model specialized in science and technology; and by the Institute of Information \& communications Technology Planning \& Evaluation (IITP) grant funded by the Korea government (MSIT) (RS-2025-02263754, Human-Centric Embodied AI Agents with Autonomous Decision-Making); by grant (25202MFDS003) from Ministry of Food and Drug Safety in 2025; by AI-BIO Research Grant through Seoul National University; Institute of Information \& communications Technology Planning \& Evaluation (IITP) grant funded by the Korea government (MSIT) (No.\ RS-2025-25442149, LG AI STAR Talent Development Program for Leading Large-Scale Generative AI Models in the Physical AI Domain).

\bibliography{anthology,custom}

\appendix

\section{Dataset Construction Details}
\label{app:dataset}

\subsection{Domain Selection}
\label{app:domains}

\textbf{Finance.} SEC 10-K reports (2023--2024) from six technology companies: Apple, Microsoft, Amazon, Meta, Tesla, and NVIDIA. These reports exhibit high redundancy due to standardized disclosure requirements and repeated regulatory language.

\textbf{Patents.} USPTO documents (2023--2024) from five technical areas: H01M* (batteries and electrochemistry), H01L* (semiconductors and solid-state devices), H04W* (wireless communication networks), A61K* (pharmaceutical preparations), G06F* (computing and data processing). We selected 10 documents per area (50 total).

\textbf{Legal.} U.S. Code (2023) covering six titles: Title 1 (General Provisions), Title 9 (Arbitration), Title 17 (Copyrights), Title 35 (Patents), Title 40 (Public Buildings, Property, and Works), Title 41 (Public Contracts). Legal codes exhibit high similarity through cross-referencing and standardized legal language.

\textbf{General-Wiki.} Wikipedia passages randomly sampled from the HotpotQA dataset, serving as a low-redundancy baseline where articles are topically distinct with minimal factual overlap.

\subsection{Implementation Details}
\label{app:impl}

\textbf{Hyperparameters.} Multi-hop depth $M \in \{1,2,3,4\}$; Atomic Information Pool Size in Question Generation $N=100$; Question Candidates Size $S=10$ per hop.

\textbf{LLM Models.} GPT-5 Nano for atomic extraction, filtering, and ranking stages. GPT-5 for question generation to encourage diverse candidate generation.

\textbf{Chunking.} Documents split into 512-token chunks using recursive text splitting with no overlap. Chunk boundaries respect sentence boundaries to maintain semantic coherence.

\textbf{Embedding.} OpenAI text-embedding-3-large (dimension=3072) for all similarity computations. Cosine similarity threshold $\tau=0.5$ for redundancy detection candidate retrieval.

\textbf{Top-$k$ Selection.} Top-3 atomic units per chunk selected after CRRF ranking based on five quality criteria (validity, completeness, specificity, clarity, questionability).

\textbf{Inference Hardware.} NVIDIA L4 GPU with fp16 precision for all embedding computations. API-based LLM calls do not require local GPU resources.

\textbf{Reproducibility.} Fixed random seed (42) for all sampling operations.

\subsection{Similarity and Redundancy Metrics}
\label{app:sim-red}

\paragraph{Similarity.}
We report \emph{Similarity (\%)} at the chunk (passage) level. Each corpus is split into non-overlapping 512-token chunks using recursive text splitting. Let $M$ be the number of chunks and $\mathbf{e}(p)$ be the \texttt{text-embedding-3-large} embedding of chunk $p$. We compute similarity as the mean cosine similarity over all \emph{distinct} chunk pairs (i.e., excluding the diagonal):
\[
\mathrm{Sim} = \mathbb{E}_{i<j}\left[\cos\!\left(\mathbf{e}(p_i), \mathbf{e}(p_j)\right)\right].
\]
Because each corpus contains roughly 3--4K chunks, we compute the full pairwise matrix. Equivalently, this is the arithmetic mean over all off-diagonal entries in the upper triangle of the $M\times M$ pairwise cosine similarity matrix. This metric is computed over chunks, not atomic units. We use chunk-level cosine similarity as a proxy for corpus-level semantic similarity because retrieval operates over chunk-sized units, and confusion among semantically similar chunks is a primary failure mode in high-overlap corpora. 

\paragraph{Redundancy.}
We report \emph{Redundancy (\%)} at the atomic level. From each chunk, we extract atomic information units, apply validity filtering, and rank the remaining units with CRRF. For tractability, we select the top-$k$ ranked valid atomic units per chunk (default $k{=}3$) to form a target set $\mathcal{T}$. Redundancy is then defined as the fraction of targets that have at least one semantically equivalent atomic unit extracted from a different chunk:
\[
\mathrm{Red} = \frac{|\{a \in \mathcal{T} : \exists\, a' \notin \mathrm{chunk}(a)\ \text{s.t.}\ a \equiv a' \}|}{|\mathcal{T}|},
\]
where $\mathrm{chunk}(a)$ denotes the source chunk of $a$ and $a \equiv a'$ denotes semantic equivalence as determined by the redundancy tracking procedure (Section~\ref{sec:method}).

\subsection{Corpus Statistics}
\label{app:corpus-stats}

\begin{table}[h]
\centering
\caption{Detailed corpus statistics across four domains.}
\label{tab:corpus-stats}
\begin{adjustbox}{width=\columnwidth}

\begin{tabular}{lrrrr}
\toprule
\textbf{Metric} & \textbf{Finance} & \textbf{Patent} & \textbf{Legal} & \textbf{General-Wiki} \\
\midrule
Chunks & 3{,}281 & 3{,}339 & 4{,}090 & 3{,}218 \\
Total atoms & 42{,}825 & 39{,}282 & 24{,}815 & 26{,}093 \\
Valid pass rate (\%) & 55.6 & 57.5 & 58.0 & 73.2 \\
Top-3 atoms & 6{,}040 & 6{,}875 & 7{,}717 & 8{,}232 \\
Unique atoms & 2{,}221 & 3{,}460 & 5{,}783 & 8{,}120 \\
Redundancy (\%) & 63.2 & 49.7 & 25.1 & 1.4 \\
Similarity (\%) & 35.1 & 29.0 & 40.7 & 8.8 \\
\midrule
Final samples & 645 & 480 & 605 & 688 \\
\bottomrule
\end{tabular}
\end{adjustbox}
\end{table}

\textbf{Valid pass rate} indicates the proportion of extracted atomic units that pass the valid information filtering. General-Wiki's higher pass rate (73.2\%) reflects cleaner Wikipedia text compared to real-world data.

\textbf{Unique atoms} are determined after redundancy tracking: atoms that do not have semantic equivalents in other chunks. The ratio of unique to top-3 atoms quantifies redundancy---Finance retains only 36.8\% unique atoms (2,221/6,040), while General-Wiki retains 98.6\% (8,120/8,232).

\subsection{Question Filtering Statistics}
\label{app:filtering}

We generated 10 candidate questions per sample (300 samples $\times$ 4 hops $\times$ 4 domains $\times$ 10 paraphrased questions = 48,000 candidates total) and applied zero-tolerance logical filtering with five criteria. Tables~\ref{tab:filtering-stats-1} and~\ref{tab:filtering-stats-2} show pass rates by domain and hop depth.

\begin{table}[t]
\centering
\caption{Question filtering pass rates for Finance and Patent domains. All values in percentages. CI = Contextual Independence; AE = Answer Exclusion; IE = Information Equivalence; QC = Question Clarity; AS = Answerability.}
\label{tab:filtering-stats-1}
\begin{adjustbox}{max width=\columnwidth}
\small
\begin{tabular}{llcccccc}
\toprule
\textbf{Domain} & \textbf{Hop} & \textbf{CI} & \textbf{AE} & \textbf{IE} & \textbf{QC} & \textbf{AS} & \textbf{Success} \\
\midrule
\multirow{4}{*}{Finance}
& 1 & 70.8 & 74.4 & 97.2 & 99.0 & 98.4 & 50.7 \\
& 2 & 90.8 & 81.1 & 65.5 & 96.9 & 95.5 & 56.7 \\
& 3 & 90.2 & 79.0 & 75.1 & 99.0 & 94.0 & 56.7 \\
& 4 & 92.0 & 68.4 & 78.0 & 98.7 & 95.8 & 51.0 \\
\midrule
\multirow{4}{*}{Patent}
& 1 & 68.8 & 63.0 & 98.3 & 98.1 & 95.8 & 42.3 \\
& 2 & 80.1 & 66.9 & 65.2 & 98.4 & 93.8 & 38.0 \\
& 3 & 93.0 & 64.7 & 70.8 & 97.5 & 93.5 & 45.0 \\
& 4 & 88.1 & 51.8 & 74.8 & 97.5 & 93.8 & 34.7 \\
\bottomrule
\end{tabular}
\end{adjustbox}
\end{table}

\begin{table}[t]
\centering
\caption{Question filtering pass rates for Legal and General-Wiki domains. All values in percentages. Abbreviations same as Table~\ref{tab:filtering-stats-1}.}
\label{tab:filtering-stats-2}
\begin{adjustbox}{max width=\columnwidth}
\small
\begin{tabular}{llcccccc}
\toprule
\textbf{Domain} & \textbf{Hop} & \textbf{CI} & \textbf{AE} & \textbf{IE} & \textbf{QC} & \textbf{AS} & \textbf{Success} \\
\midrule
\multirow{4}{*}{Legal}
& 1 & 74.9 & 69.8 & 98.1 & 97.7 & 98.0 & 51.3 \\
& 2 & 86.9 & 75.0 & 69.7 & 98.3 & 94.9 & 53.0 \\
& 3 & 92.5 & 71.6 & 69.0 & 98.9 & 94.3 & 50.7 \\
& 4 & 92.4 & 63.4 & 77.2 & 97.6 & 92.4 & 46.7 \\
\midrule
\multirow{4}{*}{General-Wiki}
& 1 & 94.7 & 80.9 & 98.8 & 95.8 & 98.1 & 71.3 \\
& 2 & 90.7 & 72.0 & 88.9 & 95.3 & 98.7 & 58.3 \\
& 3 & 89.8 & 63.6 & 88.4 & 96.5 & 97.4 & 53.0 \\
& 4 & 86.2 & 56.9 & 85.9 & 92.2 & 98.6 & 46.7 \\
\bottomrule
\end{tabular}
\end{adjustbox}
\end{table}

\paragraph{Key Observations.}

\textbf{Information Equivalence} exhibits a characteristic U-shaped pattern: high at 1-hop (97--98\%), dropping sharply at 2-hop (65--89\%), then recovering at 3--4 hops (69--88\%). This reflects LLM generation behavior: at 1-hop, matching question scope to a single atomic unit is straightforward; at 2-hop, models frequently under-constrain (requiring only one unit) or over-constrain (requiring additional unstated facts); at higher hops, more careful candidate construction improves alignment. General-Wiki maintains consistently higher pass rates (86--99\%) than enterprise domains (65--98\%), likely due to cleaner Wikipedia text and well-structured reasoning chains.

\textbf{Answer Exclusion} degrades monotonically with hop depth across all domains, dropping from 63--81\% at 1-hop to 52--69\% at 4-hop. Multi-hop questions increasingly embed intermediate reasoning steps or partial answers within the question text itself, violating the requirement that questions must not reveal answer content.

\textbf{Contextual Independence} improves from 1-hop (69--95\%) to higher hops (86--93\%), except in General-Wiki which starts high and remains stable. At 1-hop, LLMs frequently generate meta-questions (``What section discusses X?'') that reference document structure; this tendency decreases at higher hops where complex reasoning demands content-focused formulation.

\textbf{Question Clarity} and \textbf{Answerability} remain consistently high (92--99\%) across all conditions, indicating that when LLMs generate questions, they typically produce well-formed and answerable queries---the primary challenge lies in precisely controlling question scope (Information Equivalence) and avoiding answer leakage (Answer Exclusion).

\paragraph{Overall Success Rates.}

Aggregating across all domains and hops: \textbf{2,418 questions retained from 4,800 sample instances (50.4\% sample-level success rate)}; each sample draws from a pool of 10 paraphrased candidates (48,000 total), and a sample is counted as retained when at least one candidate passes all five criteria. This reflects our conservative filtering approach where any single criterion failure results in rejection.

Domain-specific patterns reveal the impact of corpus characteristics on question generation difficulty. General-Wiki achieves the highest success rates (71.3\% at 1-hop, declining to 46.7\% at 4-hop), benefiting from cleaner Wikipedia text and well-structured reasoning chains. Enterprise domains show lower and more variable success: Finance maintains relatively stable rates (50.7--56.7\%), Legal shows moderate performance (46.7--53.0\%), while Patent exhibits the lowest success (34.7--45.0\%), likely due to dense technical terminology and complex domain-specific constructs that challenge precise question formulation.

The 50.4\% overall sample-level success rate---retaining half of the sample instances---reflects our design priority: dataset quality over quantity. By rejecting any candidate failing even one of five criteria, we ensure RedQA contains only logically sound, precisely scoped questions suitable for valid retrieval evaluation.

\subsection{Cost Analysis}
\label{app:cost}

\begin{table}[h]
\centering
\caption{Construction cost (USD) by domain and pipeline stage, based on OpenAI API pricing.}
\label{tab:cost}
\small
\begin{adjustbox}{width=\columnwidth}
\begin{tabular}{lrrrrr}
\toprule
\textbf{Stage} & \textbf{Finance} & \textbf{Patent} & \textbf{Legal} & \textbf{General-Wiki} & \textbf{Total} \\
\midrule
Atomic extraction & 6.88 & 6.51 & 7.86 & 4.92 & 26.17 \\
Valid selection & 18.97 & 18.66 & 21.21 & 18.11 & 76.95 \\
Embedding & 0.07 & 0.01 & 0.06 & 0.04 & 0.18 \\
Redundancy tracking & 30.93 & 32.14 & 30.97 & 2.69 & 96.73 \\
Question generation & 110.03 & 112.31 & 110.56 & 127.17 & 460.07 \\
\midrule
\textbf{Total per domain} & 166.88 & 169.63 & 170.66 & 152.93 & \textbf{660.10} \\
\bottomrule
\end{tabular}
\end{adjustbox}
\end{table}

\textbf{Question generation} accounts for 69.7\% of total cost (\$460.07 / \$660.10), driven by (i) generating 10 candidate questions per sample using GPT-5, and (ii) filtering each candidate against five strict logical criteria.

\textbf{Redundancy tracking} is more expensive in high-redundancy domains (Finance: \$30.93, Patent: \$32.14) than in General-Wiki (\$2.69) because the number of equivalence checks scales with redundancy: Finance's 6,040 top-3 atoms compress to only 2,221 unique atoms, requiring 3,819 LLM-based equivalence verifications.

\textbf{Embedding costs} are negligible (\$0.18 total) because text-embedding-3-large is efficient and only requires one embedding per atomic unit.

\textbf{Per-sample cost} averages \$0.26 (Finance), \$0.35 (Patent), \$0.28 (Legal), and \$0.22 (General-Wiki), making RARE practical for constructing domain-specific benchmarks at scale.

\section{Atomic Information Examples from RedQA}
\label{app:atomic-examples}

\paragraph{Finance.}
\begin{itemize}[leftmargin=*,noitemsep,topsep=2pt]
\item \textit{``As of December 31, 2023, Tesla's sales-type lease receivables maturing after 2028 totaled \$2 million.''}
\item \textit{``Meta's total goodwill from acquisitions in 2022 was \$1,137 million.''}
\item \textit{``Amazon's long-term debt was \$52,623 million as of December 31, 2024.''}
\item \textit{``Apple's Vision Pro is expected to be available in early calendar year 2024.''}
\item \textit{``Apple manages its business primarily on a geographic basis.''}
\item \textit{``NVIDIA purchases memory from SK Hynix Inc.''}
\item \textit{``Meta's board of directors will consider continued capital availability, market conditions, and applicable laws and agreements when deciding whether to declare cash dividends.''}
\end{itemize}

\paragraph{Legal.}
\begin{itemize}[leftmargin=*,noitemsep,topsep=2pt]
\item \textit{``For competitive basis purchases, the notice must include the basis on which the selection will be made.''}
\item \textit{``International applications shall be processed by the United States Patent and Trademark Office when acting as a Receiving Office, International Searching Authority, or International Preliminary Examining Authority.''}
\item \textit{``Executive Order 13658, issued on February 12, 2014, established a minimum wage for federal contractors.''}
\item \textit{``An officer of the mechanical licensing collective may not simultaneously be an employee of any board member.''}
\item \textit{``Each joint owner of a patent may make, use, offer to sell, or sell the patented invention.''}
\item \textit{``Made in America Laws require or provide a preference for the purchase or acquisition of goods, products, or materials produced in the United States.''}
\end{itemize}

\paragraph{Patent.}
\begin{itemize}[leftmargin=*,noitemsep,topsep=2pt]
\item \textit{``The cancer stem cell surface marker-binding agent can be labeled with a paramagnetic label.''}
\item \textit{``Aluminum is a typical material for the positive current collector in lithium cells.''}
\item \textit{``An aluminum nitride film can be used as the protection insulating layer.''}
\item \textit{``The CMP composition contains 0.01 wt\% to 1 wt\% of at least one non-ionic surfactant.''}
\item \textit{``The encoder unit is configured to receive content from a plurality of content providers.''}
\item \textit{``CD140a is expressed on fetal human oligodendrocyte progenitor cells.''}
\item \textit{``In a wireless transmit/receive unit (WTRU), the transceiver is configured to receive a second command that reactivates the second cell.''}
\end{itemize}

\section{Qualitative Examples}
\label{app:examples}

We provide representative multi-hop questions from RedQA to illustrate the reasoning complexity and evidence requirements at each hop depth.

\subsection{1-Hop Questions}

\textbf{Example 1.1}
\begin{description}[leftmargin=1em, labelwidth=0pt, itemsep=3pt]
    \item[\textit{Question:}] When Meta Platforms, Inc. prepares its consolidated financial statements, whose accounts are included in scope?
    \item[\textit{Answer:}] The accounts of Meta Platforms, Inc. itself, its subsidiaries in which Meta has a controlling financial interest, and any variable interest entities for which Meta is deemed the primary beneficiary.
    \item[\textit{Evidence:}] The consolidated financial statements include the accounts of Meta Platforms, Inc., its subsidiaries where Meta has controlling financial interests, and any variable interest entities for which Meta is deemed the primary beneficiary.
\end{description}

\vspace{0.2cm}

\textbf{Example 1.2}
\begin{description}[leftmargin=1em, labelwidth=0pt, itemsep=3pt]
    \item[\textit{Question:}] As part of a strategic review of large-model deployments, how long did it take to train a GPT with about 1.8 trillion parameters, and which hardware was used?
    \item[\textit{Answer:}] Duration: about 3--5 months; Hardware: 25,000 NVIDIA Ampere GPUs.
    \item[\textit{Evidence:}] Training a 1.8 trillion-parameter GPT model took about three to five months using 25,000 NVIDIA Ampere GPUs.
\end{description}

\vspace{0.2cm}

\subsection{2-Hop Questions}

\textbf{Example 2.1}
\begin{description}[leftmargin=1em, labelwidth=0pt, itemsep=3pt]
    \item[\textit{Question:}] Who is responsible for establishing and maintaining internal control over financial reporting, and which top executives formulate or supervise that framework?
    \item[\textit{Answer:}] Management is responsible for establishing and maintaining ICFR; the framework is designed or supervised by the Chief Executive Officer and the Chief Financial Officer.
    \item[\textit{Evidence:}] (1) Management is responsible for establishing and maintaining adequate internal control over financial reporting. (2) Internal control over financial reporting is a process designed by, or under the supervision of, our Chief Executive Officer and Chief Financial Officer.
\end{description}

\vspace{0.2cm}

\textbf{Example 2.2}
\begin{description}[leftmargin=1em, labelwidth=0pt, itemsep=3pt]
    \item[\textit{Question:}] What material is commonly used for the layer applied to the main portion of the anode current collector in a lithium-ion battery?
    \item[\textit{Answer:}] Graphitic carbon (carbon-based negative electrode material).
    \item[\textit{Evidence:}] (1) Graphitic carbon is commonly used as the negative-electrode material in lithium-ion batteries. (2) The main region of the anode current collector is loaded with a layer of negative electrode material.
\end{description}

\vspace{0.2cm}

\subsection{3-Hop Questions}

\textbf{Example 3.1}
\begin{description}[leftmargin=1em, labelwidth=0pt, itemsep=3pt]
    \item[\textit{Question:}] Identify the attestation provider, the auditing standards that mandate reasonable assurance for the financial statements audit, and the exact internal control framework applied in auditing internal control over financial reporting.
    \item[\textit{Answer:}] Attestation provider: registered public accounting firm; Standards: PCAOB; ICFR framework: COSO Internal Control~---~Integrated Framework (2013).
    \item[\textit{Evidence:}] (1) PCAOB standards require that we plan and perform the audit to obtain reasonable assurance about whether the financial statements are free of material misstatement due to error or fraud. (2) The registrant has filed a report on and attestation to its management's assessment of the effectiveness of its internal control over financial reporting under Section 404(b) of the Sarbanes-Oxley Act by the registered public accounting firm that prepared or issued its audit report. (3) The Company's internal control over financial reporting as of January 28, 2024 was audited based on criteria established in Internal Control~---~Integrated Framework (2013) issued by COSO.
\end{description}

\vspace{0.2cm}

\textbf{Example 3.2}
\begin{description}[leftmargin=1em, labelwidth=0pt, itemsep=3pt]
    \item[\textit{Question:}] For a quarterly market outlook, how could the Ethereum 2.0 upgrade affect demand for NVIDIA's latest graphics cards across its expanded customer base, given the possibility that higher resale of GPUs might dampen new-card purchases?
    \item[\textit{Answer:}] Ethereum 2.0 is likely to reduce Ethereum mining demand and increase aftermarket GPU supply, which could depress retail prices and reduce demand for NVIDIA's new GPUs. Because NVIDIA now serves broader markets beyond gaming, the aftermarket-driven pressure is concentrated in the gaming segment rather than determining overall company demand.
    \item[\textit{Evidence:}] (1) Changes to cryptocurrency standards and processes, including the Ethereum 2.0 merge, may create increased aftermarket sales of our GPUs. (2) NVIDIA's business has evolved from gaming products to broader markets. (3) Increased aftermarket sales of our GPUs could reduce demand for our new GPUs.
\end{description}

\vspace{0.2cm}

\subsection{4-Hop Questions}

\textbf{Example 4.1}
\begin{description}[leftmargin=1em, labelwidth=0pt, itemsep=3pt]
    \item[\textit{Question:}] When regulations and platform policies restrict ad data processing, what practice does Meta use to match accounts, which mobile OS change exacerbated limits on ad effectiveness, which specific legal frameworks are cited as restricting the use of signals, and what EU decision was adopted on July 10, 2023 regarding the United States?
    \item[\textit{Answer:}] Meta uses data signals such as device information to match user accounts. The mobile OS change that exacerbated limits on ad effectiveness was Apple's iOS updates. The regulatory frameworks cited are GDPR, the ePrivacy Directive, the European Digital Services Act (DSA), the Digital Markets Act (DMA), and U.S. state privacy laws including the California Consumer Privacy Act as amended by the California Privacy Rights Act. On July 10, 2023, the European Commission adopted an adequacy decision in relation to the United States under the EU-U.S. Data Privacy Framework.
    \item[\textit{Evidence:}] (1) Regulatory developments including GDPR, evolving interpretations by the Court of Justice of the European Union, ePrivacy Directive, European Digital Services Act, Digital Markets Act, and U.S. state privacy laws including the California Consumer Privacy Act as amended by the California Privacy Rights Act have impacted Meta's ability to use data signals in ad products. (2) We use device information as data signals to match user accounts. (3) Apple released changes to iOS that limit Meta's ability to target and measure ads effectively. (4) On July 10, 2023, the European Commission adopted an adequacy decision in relation to the United States; the adequacy decision is pursuant to the EU-U.S. Data Privacy Framework (EU-U.S. DPF), which replaces two prior adequacy frameworks invalidated by the CJEU.
\end{description}

\vspace{0.2cm}

\textbf{Example 4.2}
\begin{description}[leftmargin=1em, labelwidth=0pt, itemsep=3pt]
  \item[\textit{Question:}] In NVIDIA's compensation disclosures, which individuals are listed as
non-CEO named executive officers, what form of equity dominates their target compensation, what is the
maximum payout as a percentage of target for their target-based awards, and how does the CEO's equity
setup compare?
  \item[\textit{Answer:}] Non-CEO NEOs: Colette M. Kress, Ajay K. Puri, Debora Shoquist, Timothy S.
Teter. Their target pay is dominated by at-risk equity in the form of PSUs (60\% PSUs, 40\% RSUs). The
maximum payout on target-based awards is 200\% of target. The CEO's equity setup is different: Jensen
Huang's target equity is 100\% at-risk PSUs (split 50-50 between MY PSUs and SY PSUs).
  \item[\textit{Evidence:}] (1) For Fiscal 2023, the Compensation Committee (CC) decided that the
largest portion of NEOs' total target pay would remain in the form of at-risk equity with
performance-based vesting; for non-CEO NEOs, the CC provided 40\% of the target equity opportunity in
the form of RSUs and 60\% in the form of PSUs. (2) The maximum payout for NEOs under target-based
awards is capped at 200\% of target. (3) For Fiscal 2024, 2023, 2022, and 2021, our non-CEO NEOs were
Colette M. Kress, Ajay K. Puri, Debora Shoquist, and Timothy S. Teter. (4) The CC concluded that 100\%
of Mr. Huang's equity grants should be at-risk and performance-based, and granted his target equity
opportunity 100\% in the form of SY PSUs and MY PSUs, evenly split between both forms of PSUs.
\end{description}

\subsection{Observations}
As hop depth increases, questions require synthesizing information from progressively more dispersed evidence passages. At 4-hop, retrievers must aggregate facts across regulatory frameworks, corporate policies, compensation structures, and specific role definitions---all while navigating high document redundancy that introduces numerous near-duplicate distractors. At 2+ hops, answers rarely reside in a single passage, mirroring real-world RAG deployments.

\section{Complete Experimental Results}
\label{app:results}

\subsection{Main Results: All Metrics}

Tables~\ref{tab:finance-full} through \ref{tab:hotpot-full} present complete results (Coverage@10, PerfRecall@10, NDCG@10, MRR) across all models, domains, and hop depths.
\begin{table}[t]
\centering
\caption{Finance: complete results by hop depth.}
\label{tab:finance-full}
\begin{adjustbox}{width=\columnwidth}
\begin{tabular}{lcccc|cccc}
\toprule
& \multicolumn{4}{c}{\textbf{Overall (645)}} & \multicolumn{4}{c}{\textbf{1-Hop (152)}} \\
\cmidrule(lr){2-5}\cmidrule(lr){6-9}
\textbf{Model} & Coverage@10 & PerfRecall@10 & NDCG@10 & MRR & Coverage@10 & PerfRecall@10 & NDCG@10 & MRR \\
\midrule
BM25 & 59.9 & 36.1 & 0.485 & 0.361 & 82.9 & 82.9 & 0.646 & 0.587 \\
OpenAI-Large & 67.7 & 42.9 & 0.563 & 0.421 & 90.1 & 90.1 & 0.713 & 0.655 \\
E5-Large & 62.8 & 38.0 & 0.520 & 0.394 & 84.9 & 84.9 & 0.673 & 0.618 \\
BGE-M3 & 64.4 & 39.8 & 0.526 & 0.393 & 83.6 & 83.6 & 0.647 & 0.587 \\
Qwen3-0.6B & 67.6 & 43.9 & 0.549 & 0.404 & 86.8 & 86.8 & 0.697 & 0.643 \\
Jina-v4 & 65.1 & 40.3 & 0.551 & 0.421 & 86.2 & 86.2 & 0.742 & 0.704 \\
Qwen3-4B & 72.4 & 48.8 & 0.594 & 0.440 & 89.5 & 89.5 & 0.760 & 0.717 \\
E5-Mistral & 69.7 & 44.8 & 0.567 & 0.427 & 88.2 & 88.2 & 0.720 & 0.669 \\
Qwen3-8B & 72.9 & 47.4 & 0.604 & 0.448 & 90.1 & 90.1 & 0.754 & 0.707 \\
\midrule
& \multicolumn{4}{c}{\textbf{2-Hop (170)}} & \multicolumn{4}{c}{\textbf{3-Hop (170)}} \\
\cmidrule(lr){2-5}\cmidrule(lr){6-9}
BM25 & 60.3 & 36.5 & 0.475 & 0.354 & 52.2 & 20.6 & 0.433 & 0.287 \\
OpenAI-Large & 69.1 & 47.6 & 0.578 & 0.438 & 60.0 & 24.7 & 0.507 & 0.332 \\
E5-Large & 62.9 & 39.4 & 0.510 & 0.392 & 56.5 & 21.8 & 0.476 & 0.321 \\
BGE-M3 & 67.1 & 44.7 & 0.543 & 0.419 & 55.5 & 22.9 & 0.464 & 0.297 \\
Qwen3-0.6B & 70.0 & 50.6 & 0.547 & 0.402 & 61.0 & 28.2 & 0.502 & 0.319 \\
Jina-v4 & 66.5 & 45.3 & 0.558 & 0.430 & 58.0 & 23.5 & 0.471 & 0.307 \\
Qwen3-4B & 76.8 & 60.6 & 0.595 & 0.435 & 66.3 & 32.9 & 0.538 & 0.346 \\
E5-Mistral & 68.8 & 49.4 & 0.541 & 0.415 & 62.9 & 28.8 & 0.518 & 0.345 \\
Qwen3-8B & 77.9 & 59.4 & 0.612 & 0.452 & 66.1 & 32.4 & 0.556 & 0.361 \\
\midrule
& \multicolumn{4}{c}{\textbf{4-Hop (153)}} \\
\cmidrule(lr){2-5}
BM25 & 45.1 & 6.5 & 0.395 & 0.227 \\
OpenAI-Large & 52.3 & 11.1 & 0.458 & 0.269 \\
E5-Large & 47.9 & 7.8 & 0.426 & 0.256 \\
BGE-M3 & 52.5 & 9.8 & 0.455 & 0.276 \\
Qwen3-0.6B & 53.3 & 11.1 & 0.454 & 0.264 \\
Jina-v4 & 50.5 & 7.8 & 0.442 & 0.256 \\
Qwen3-4B & 57.5 & 13.1 & 0.487 & 0.274 \\
E5-Mistral & 59.8 & 14.4 & 0.500 & 0.290 \\
Qwen3-8B & 57.8 & 8.5 & 0.498 & 0.282 \\
\bottomrule
\end{tabular}
\end{adjustbox}
\end{table}

\begin{table}[t]
\centering
\caption{Patent: complete results by hop depth.}
\label{tab:patent-full}
\begin{adjustbox}{width=\columnwidth}
\begin{tabular}{lcccc|cccc}
\toprule
& \multicolumn{4}{c}{\textbf{Overall (480)}} & \multicolumn{4}{c}{\textbf{1-Hop (127)}} \\
\cmidrule(lr){2-5}\cmidrule(lr){6-9}
\textbf{Model} & Coverage@10 & PerfRecall@10 & NDCG@10 & MRR & Coverage@10 & PerfRecall@10 & NDCG@10 & MRR \\
\midrule
BM25 & 76.4 & 55.6 & 0.632 & 0.488 & 88.2 & 88.2 & 0.712 & 0.657 \\
OpenAI-Large & 82.0 & 61.5 & 0.665 & 0.497 & 92.1 & 92.1 & 0.732 & 0.670 \\
E5-Large & 80.2 & 60.6 & 0.649 & 0.482 & 90.6 & 90.6 & 0.693 & 0.624 \\
BGE-M3 & 81.8 & 59.6 & 0.674 & 0.510 & 93.7 & 93.7 & 0.751 & 0.690 \\
Qwen3-0.6B & 81.5 & 60.2 & 0.680 & 0.515 & 91.3 & 91.3 & 0.730 & 0.670 \\
Jina-v4 & 80.1 & 59.6 & 0.666 & 0.510 & 89.0 & 89.0 & 0.702 & 0.643 \\
Qwen3-4B & 83.1 & 62.5 & 0.708 & 0.551 & 94.5 & 94.5 & 0.769 & 0.713 \\
E5-Mistral & 81.7 & 62.7 & 0.657 & 0.486 & 92.1 & 92.1 & 0.706 & 0.636 \\
Qwen3-8B & 84.0 & 63.1 & 0.714 & 0.546 & 94.5 & 94.5 & 0.775 & 0.721 \\
\midrule
& \multicolumn{4}{c}{\textbf{2-Hop (114)}} & \multicolumn{4}{c}{\textbf{3-Hop (135)}} \\
\cmidrule(lr){2-5}\cmidrule(lr){6-9}
BM25 & 77.2 & 61.4 & 0.616 & 0.490 & 74.8 & 44.4 & 0.632 & 0.430 \\
OpenAI-Large & 84.2 & 70.2 & 0.656 & 0.492 & 79.0 & 50.4 & 0.660 & 0.439 \\
E5-Large & 77.2 & 64.0 & 0.606 & 0.455 & 78.5 & 52.6 & 0.661 & 0.436 \\
BGE-M3 & 82.5 & 65.8 & 0.639 & 0.482 & 78.5 & 49.6 & 0.670 & 0.445 \\
Qwen3-0.6B & 81.6 & 67.5 & 0.659 & 0.509 & 79.0 & 46.7 & 0.683 & 0.456 \\
Jina-v4 & 78.5 & 63.2 & 0.623 & 0.482 & 79.8 & 51.1 & 0.693 & 0.479 \\
Qwen3-4B & 83.3 & 67.5 & 0.687 & 0.542 & 80.2 & 51.9 & 0.706 & 0.499 \\
E5-Mistral & 81.1 & 67.5 & 0.633 & 0.465 & 80.7 & 57.0 & 0.669 & 0.442 \\
Qwen3-8B & 85.1 & 71.1 & 0.710 & 0.553 & 80.5 & 54.1 & 0.700 & 0.472 \\
\midrule
& \multicolumn{4}{c}{\textbf{4-Hop (104)}} \\
\cmidrule(lr){2-5}
BM25 & 63.0 & 24.0 & 0.551 & 0.354 \\
OpenAI-Large & 70.9 & 28.8 & 0.599 & 0.369 \\
E5-Large & 73.1 & 30.8 & 0.628 & 0.399 \\
BGE-M3 & 70.9 & 24.0 & 0.622 & 0.407 \\
Qwen3-0.6B & 72.6 & 31.7 & 0.639 & 0.406 \\
Jina-v4 & 71.4 & 30.8 & 0.634 & 0.418 \\
Qwen3-4B & 72.6 & 31.7 & 0.657 & 0.430 \\
E5-Mistral & 70.7 & 28.8 & 0.606 & 0.385 \\
Qwen3-8B & 74.8 & 27.9 & 0.662 & 0.421 \\
\bottomrule
\end{tabular}
\end{adjustbox}
\end{table}

\begin{table}[t]
\centering
\caption{Legal: complete results by hop depth.}
\label{tab:legal-full}
\begin{adjustbox}{width=\columnwidth}
\begin{tabular}{lcccc|cccc}
\toprule
& \multicolumn{4}{c}{\textbf{Overall (605)}} & \multicolumn{4}{c}{\textbf{1-Hop (154)}} \\
\cmidrule(lr){2-5}\cmidrule(lr){6-9}
\textbf{Model} & Coverage@10 & PerfRecall@10 & NDCG@10 & MRR & Coverage@10 & PerfRecall@10 & NDCG@10 & MRR \\
\midrule
BM25 & 49.1 & 27.8 & 0.402 & 0.299 & 68.8 & 68.8 & 0.569 & 0.531 \\
OpenAI-Large & 61.2 & 37.4 & 0.489 & 0.352 & 79.2 & 79.2 & 0.602 & 0.542 \\
E5-Large & 52.6 & 30.9 & 0.408 & 0.290 & 64.9 & 64.9 & 0.487 & 0.436 \\
BGE-M3 & 54.5 & 32.4 & 0.429 & 0.306 & 70.1 & 70.1 & 0.517 & 0.459 \\
Qwen3-0.6B & 60.3 & 35.4 & 0.466 & 0.331 & 77.3 & 77.3 & 0.560 & 0.494 \\
Jina-v4 & 59.6 & 34.5 & 0.474 & 0.340 & 76.6 & 76.6 & 0.574 & 0.513 \\
Qwen3-4B & 65.1 & 39.7 & 0.538 & 0.391 & 83.1 & 83.1 & 0.642 & 0.581 \\
E5-Mistral & 56.9 & 34.0 & 0.433 & 0.310 & 71.4 & 71.4 & 0.528 & 0.469 \\
Qwen3-8B & 67.2 & 41.5 & 0.571 & 0.425 & 85.7 & 85.7 & 0.704 & 0.656 \\
\midrule
& \multicolumn{4}{c}{\textbf{2-Hop (159)}} & \multicolumn{4}{c}{\textbf{3-Hop (152)}} \\
\cmidrule(lr){2-5}\cmidrule(lr){6-9}
BM25 & 51.9 & 29.6 & 0.397 & 0.291 & 38.6 & 8.6 & 0.329 & 0.205 \\
OpenAI-Large & 64.5 & 44.7 & 0.510 & 0.368 & 56.8 & 19.1 & 0.470 & 0.292 \\
E5-Large & 57.5 & 36.5 & 0.435 & 0.313 & 49.3 & 17.8 & 0.399 & 0.244 \\
BGE-M3 & 59.7 & 40.9 & 0.455 & 0.328 & 48.7 & 14.5 & 0.429 & 0.275 \\
Qwen3-0.6B & 62.9 & 39.6 & 0.463 & 0.331 & 55.3 & 18.4 & 0.465 & 0.294 \\
Jina-v4 & 59.4 & 39.0 & 0.459 & 0.328 & 58.3 & 18.4 & 0.489 & 0.314 \\
Qwen3-4B & 66.4 & 46.5 & 0.535 & 0.401 & 59.9 & 21.7 & 0.523 & 0.331 \\
E5-Mistral & 62.6 & 40.3 & 0.478 & 0.353 & 53.3 & 17.1 & 0.410 & 0.245 \\
Qwen3-8B & 68.2 & 46.5 & 0.565 & 0.437 & 64.0 & 25.0 & 0.553 & 0.348 \\
\midrule
& \multicolumn{4}{c}{\textbf{4-Hop (140)}} \\
\cmidrule(lr){2-5}
BM25 & 35.5 & 1.4 & 0.306 & 0.156 \\
OpenAI-Large & 42.3 & 2.9 & 0.361 & 0.193 \\
E5-Large & 36.8 & 1.4 & 0.301 & 0.154 \\
BGE-M3 & 37.7 & 0.7 & 0.302 & 0.149 \\
Qwen3-0.6B & 44.1 & 2.9 & 0.369 & 0.191 \\
Jina-v4 & 42.3 & 0.7 & 0.365 & 0.193 \\
Qwen3-4B & 49.6 & 3.6 & 0.445 & 0.237 \\
E5-Mistral & 38.2 & 4.3 & 0.305 & 0.155 \\
Qwen3-8B & 48.9 & 5.0 & 0.450 & 0.242 \\
\bottomrule
\end{tabular}
\end{adjustbox}
\end{table}

\begin{table}[t]
\centering
\caption{General-Wiki: complete results by hop depth.}
\label{tab:hotpot-full}
\begin{adjustbox}{width=\columnwidth}
\begin{tabular}{lcccc|cccc}
\toprule
& \multicolumn{4}{c}{\textbf{Overall (688)}} & \multicolumn{4}{c}{\textbf{1-Hop (214)}} \\
\cmidrule(lr){2-5}\cmidrule(lr){6-9}
\textbf{Model} & Coverage@10 & PerfRecall@10 & NDCG@10 & MRR & Coverage@10 & PerfRecall@10 & NDCG@10 & MRR \\
\midrule
BM25 & 64.6 & 44.2 & 0.618 & 0.486 & 86.9 & 86.9 & 0.805 & 0.784 \\
OpenAI-Large & 92.8 & 86.5 & 0.922 & 0.702 & 100.0 & 100.0 & 0.998 & 0.998 \\
E5-Large & 91.7 & 84.0 & 0.908 & 0.693 & 100.0 & 100.0 & 0.991 & 0.988 \\
BGE-M3 & 90.6 & 82.3 & 0.896 & 0.686 & 100.0 & 100.0 & 0.991 & 0.988 \\
Qwen3-0.6B & 90.0 & 78.8 & 0.888 & 0.680 & 100.0 & 100.0 & 0.988 & 0.984 \\
Jina-v4 & 92.0 & 85.0 & 0.913 & 0.695 & 100.0 & 100.0 & 0.995 & 0.993 \\
Qwen3-4B & 94.0 & 89.8 & 0.929 & 0.705 & 100.0 & 100.0 & 0.991 & 0.988 \\
E5-Mistral & 93.5 & 88.7 & 0.923 & 0.702 & 100.0 & 100.0 & 0.994 & 0.992 \\
Qwen3-8B & 93.6 & 88.7 & 0.926 & 0.703 & 100.0 & 100.0 & 0.990 & 0.986 \\
\midrule
& \multicolumn{4}{c}{\textbf{2-Hop (175)}} & \multicolumn{4}{c}{\textbf{3-Hop (159)}} \\
\cmidrule(lr){2-5}\cmidrule(lr){6-9}
BM25 & 70.9 & 51.4 & 0.675 & 0.506 & 49.3 & 13.8 & 0.480 & 0.293 \\
OpenAI-Large & 95.1 & 92.0 & 0.939 & 0.706 & 89.5 & 83.0 & 0.889 & 0.543 \\
E5-Large & 95.4 & 93.7 & 0.935 & 0.706 & 86.6 & 76.1 & 0.854 & 0.520 \\
BGE-M3 & 96.0 & 94.3 & 0.936 & 0.701 & 85.1 & 75.5 & 0.845 & 0.516 \\
Qwen3-0.6B & 94.0 & 90.3 & 0.920 & 0.690 & 87.2 & 75.5 & 0.852 & 0.517 \\
Jina-v4 & 93.7 & 90.3 & 0.931 & 0.699 & 88.7 & 79.9 & 0.876 & 0.534 \\
Qwen3-4B & 96.9 & 95.4 & 0.952 & 0.715 & 90.6 & 84.3 & 0.896 & 0.545 \\
E5-Mistral & 96.0 & 94.3 & 0.938 & 0.708 & 90.6 & 85.5 & 0.890 & 0.540 \\
Qwen3-8B & 96.9 & 96.0 & 0.950 & 0.714 & 90.1 & 84.9 & 0.889 & 0.541 \\
\midrule
& \multicolumn{4}{c}{\textbf{4-Hop (140)}} \\
\cmidrule(lr){2-5}
BM25 & 40.0 & 4.3 & 0.418 & 0.226 \\
OpenAI-Large & 82.5 & 62.9 & 0.820 & 0.428 \\
E5-Large & 80.0 & 56.4 & 0.805 & 0.423 \\
BGE-M3 & 75.7 & 47.9 & 0.759 & 0.399 \\
Qwen3-0.6B & 72.7 & 35.7 & 0.734 & 0.388 \\
Jina-v4 & 81.4 & 61.4 & 0.805 & 0.419 \\
Qwen3-4B & 85.0 & 73.6 & 0.844 & 0.440 \\
E5-Mistral & 83.9 & 67.9 & 0.831 & 0.432 \\
Qwen3-8B & 83.6 & 66.4 & 0.839 & 0.440 \\
\bottomrule
\end{tabular}
\end{adjustbox}
\end{table}

\section{CRRF Ablation Study}
\label{app:CRRF}

We evaluate whether separate criterion evaluation combined with rank fusion (CRRF) improves ranking quality over joint prompting or confidence-based aggregation. This ablation isolates the contributions of (1) prompting strategy and (2) aggregation method.

\subsection{Experimental Setup}

\paragraph{Task.} We rank evidence sentences from HotpotQA passages based on five quality criteria used in RARE's Valid Information Selection stage: validity, completeness, specificity, clarity, and questionability. Each method ranks 100 candidate sentences, and we measure NDCG@3, MRR, and Top-1 accuracy against human judgments. Since HotpotQA passages are shorter than those typically found in industry settings, we combined the train and dev sets and selected the top 100 passages by token length, resulting in an average of 350 tokens---closely matching RedQA's average passage length of 356.95 tokens.

\paragraph{Prompting Strategies.}
\begin{itemize}[leftmargin=*,noitemsep,topsep=3pt]
    \item \textbf{Vanilla}: Single holistic prompt asking to rank sentences by overall quality without explicit criteria enumeration.
    \item \textbf{Combined}: Single prompt listing all five criteria and asking the LLM to jointly evaluate and rank sentences considering all criteria simultaneously.
    \item \textbf{Separate}: Five independent prompts, each evaluating one criterion. Each prompt produces a per-criterion ranking without considering other criteria.
\end{itemize}

\paragraph{Aggregation Methods.}
\begin{itemize}[leftmargin=*,noitemsep,topsep=3pt]
    \item \textbf{Base}: Arithmetic mean of raw LLM confidence scores across criteria. For Vanilla (single criterion), this is the raw score.
    \item \textbf{MinMax}: Min-max normalization of scores to $[0,1]$ before averaging, addressing scale mismatches between criteria.
    \item \textbf{RRF} (Reciprocal Rank Fusion): Rank-based fusion $s(x) = \sum_{i=1}^{N} \frac{1}{\text{rank}_i(x)}$, discarding confidence magnitudes entirely and relying only on ordinal preferences.
\end{itemize}

We run each configuration 5 times and report mean values.

\subsection{Complete Ablation Results}

Tables~\ref{tab:CRRF-nano} and~\ref{tab:CRRF-gpt5} present results for GPT-5 Nano and GPT-5 (5-run average).

\begin{table}[t]
\centering
\caption{CRRF ablation on GPT-5 Nano (5-run avg). Separate+RRF achieves substantial gains over all baselines.}
\label{tab:CRRF-nano}
\small
\begin{tabular}{llccc}
\toprule
\textbf{Prompt} & \textbf{Agg} & \textbf{NDCG@3} & \textbf{MRR} & \textbf{Top-1} \\
\midrule
Vanilla & Base & 0.352 & 0.401 & 22.00 \\
\midrule
Combined & Base & 0.325 & 0.371 & 16.20 \\
Combined & MinMax & 0.320 & 0.376 & 17.60 \\
Combined & RRF & 0.419 & 0.452 & 26.20 \\
\midrule
Separate & Base & 0.376 & 0.423 & 23.60 \\
Separate & MinMax & 0.391 & 0.435 & 25.00 \\
Separate & RRF & \textbf{0.463} & \textbf{0.494} & \textbf{33.00} \\
\bottomrule
\end{tabular}
\end{table}

\begin{table}[t]
\centering
\caption{CRRF ablation on GPT-5 (5-run avg). Gains persist with stronger models, confirming fundamental advantages of decomposed reasoning.}
\label{tab:CRRF-gpt5}
\small
\begin{tabular}{llccc}
\toprule
\textbf{Prompt} & \textbf{Agg} & \textbf{NDCG@3} & \textbf{MRR} & \textbf{Top-1} \\
\midrule
Vanilla & Base & 0.395 & 0.447 & 27.80 \\
\midrule
Combined & Base & 0.385 & 0.429 & 24.00 \\
Combined & MinMax & 0.395 & 0.434 & 23.60 \\
Combined & RRF & 0.454 & 0.490 & 32.20 \\
\midrule
Separate & Base & 0.400 & 0.452 & 28.40 \\
Separate & MinMax & 0.424 & 0.463 & 28.20 \\
Separate & RRF & \textbf{0.467} & \textbf{0.497} & \textbf{33.20} \\
\bottomrule
\end{tabular}
\end{table}

\subsection{Stability Analysis}

\begin{table}[t]
\centering
\caption{Cross-run variability (coefficient of variation in \%) across 5 runs. Lower values indicate more consistent behavior.}
\label{tab:CRRF-stability}
\small
\begin{adjustbox}{max width=\columnwidth}
\begin{tabular}{llcccc}
\toprule
 &  & \multicolumn{2}{c}{\textbf{GPT-5 Nano}} & \multicolumn{2}{c}{\textbf{GPT-5}} \\
\cmidrule(lr){3-4} \cmidrule(lr){5-6}
\textbf{Prompt} & \textbf{Agg} & \textbf{NDCG@3} & \textbf{MRR} & \textbf{NDCG@3} & \textbf{MRR} \\
\midrule
Vanilla & Base & 5.2 & 5.1 & 2.7 & 2.0 \\
\midrule
Combined & Base & 5.8 & 4.0 & 4.5 & 3.0 \\
Combined & MinMax & 6.6 & 5.4 & 3.7 & 2.3 \\
Combined & RRF & 1.7 & 1.4 & 4.3 & 2.8 \\
\midrule
Separate & Base & 4.1 & 2.8 & 3.1 & 1.2 \\
Separate & MinMax & 3.3 & 2.8 & \textbf{0.9} & \textbf{0.5} \\
Separate & RRF & \textbf{3.2} & \textbf{1.7} & 1.7 & 0.8 \\
\bottomrule
\end{tabular}
\end{adjustbox}
\end{table}

Beyond mean performance, stability across runs is critical for production deployment. Table~\ref{tab:CRRF-stability} shows coefficient of variation (CV = std/mean $\times$ 100\%) across 5 runs.

\paragraph{Separate Prompting Reduces Variance.}

Comparing prompting strategies within the same aggregation method reveals consistent patterns:

\textbf{Separate vs. Combined (Base):} Separate reduces CV from 5.8\% to 4.1\% (Nano) and from 4.5\% to 3.1\% (GPT-5) for NDCG@3. Independent criterion evaluation produces more stable rankings because variations in one criterion do not cascade to affect judgments of other criteria.

\textbf{Separate vs. Vanilla:} Separate+Base (4.1\% CV) shows lower variance than Vanilla+Base (5.2\%) on Nano, suggesting that explicit criterion decomposition---even without rank fusion---improves consistency by providing structured evaluation scaffolding.

\paragraph{RRF Dramatically Stabilizes Combined Prompting.}

The most striking pattern emerges in Combined prompting: RRF reduces NDCG@3 CV from 6.6\% (MinMax) to 1.7\%---a 3.9$\times$ improvement. This shows that rank-based fusion acts as a powerful variance-reduction mechanism even when individual judgments are unstable. By aggregating ordinal preferences from multiple criteria rather than relying on single-criterion cardinal scores, RRF introduces implicit ensembling that smooths out noise.

\paragraph{Model-Specific Stability Patterns.}

GPT-5 Nano shows a clear stability ranking: Separate+RRF (3.2\%) < Separate+MinMax (3.3\%) < Separate+Base (4.1\%), with all three outperforming Combined+MinMax (6.6\%). The minimal gap between RRF and MinMax (0.1 percentage point) reflects the fundamental stability of separate prompting, where aggregation method has less impact.

GPT-5 exhibits different behavior: Separate+MinMax achieves the lowest CV (0.9\% NDCG@3, 0.5\% MRR), outperforming even Separate+RRF (1.7\%, 0.8\%). This suggests that stronger models produce more calibrated confidence scores where normalization-based aggregation becomes viable. However, RRF maintains the highest absolute performance (0.467 vs. 0.424 NDCG@3), demonstrating a stability-performance tradeoff.

\paragraph{MRR vs. NDCG@3 Stability.}

MRR consistently shows lower CV than NDCG@3 across all configurations (e.g., Separate+RRF: 1.7\% vs. 3.2\% on Nano). This occurs because MRR focuses on the rank of the single highest-quality item, which is typically more stable than the aggregate quality of the top-3 items captured by NDCG@3.

\subsection{Implications for RARE}

These results directly motivate RARE's design choices:

\textbf{(1) Decompose complex judgments.} Rather than asking LLMs to rank atomic information or questions by considering all quality criteria simultaneously, RARE evaluates each criterion through independent prompts. This reduces per-call complexity and improves both reliability and consistency.

\textbf{(2) Fuse via ranks, not scores.} RARE aggregates criterion-specific rankings using RRF rather than averaging confidence scores. While Separate+MinMax achieves comparable or even superior stability on stronger models (0.9\% CV on GPT-5), RRF maintains substantially higher absolute performance (0.467 vs. 0.424 NDCG@3), making it the preferred choice when optimizing for quality rather than variance minimization alone.

\textbf{(3) Prioritize stability for benchmark construction.} In dataset generation, consistency is as important as peak performance. CRRF's low cross-run variance (1.7--3.2\% CV for NDCG@3) ensures that dataset quality remains stable across construction runs, avoiding the need for extensive manual post-hoc filtering that would arise from unstable generation. The 6.6\% CV of Combined+MinMax would introduce unacceptable variability, potentially requiring multiple regeneration attempts to achieve acceptable quality.

By adopting CRRF throughout the pipeline---for both atomic information ranking (5 criteria) and question ranking (4 criteria)---RARE achieves measurably higher quality and greater consistency than alternative multi-criterion judgment strategies.

\section{Prompt Design and Examples}
\label{app:prompts}

This section describes the prompt design for each stage of the RARE pipeline with representative examples of good and bad outputs.

\subsection{Atomic Information Extraction}
\label{app:prompt-atomic}

\textbf{Prompt Design.} The prompt instructs the LLM to extract atomic information units that will serve as ground truth answers for RAG dataset generation. Each unit must satisfy three critical criteria: (1) \textbf{Atomicity}---contains exactly one indivisible factual claim that cannot be meaningfully split, (2) \textbf{Validity}---provides substantively useful knowledge that supports real-world queries rather than trivial metadata, and (3) \textbf{Unambiguity}---completely self-contained with all entities explicitly named and no vague references. The prompt requires step-by-step reasoning for each criterion and outputs structured JSON.

\paragraph{Good Examples.}
\begin{itemize}
\item \textit{``Tesla's headquarters is located in Austin, Texas.''} (single location fact, unambiguous)
\item \textit{``The United States Declaration of Independence was adopted in 1776.''} (one historical fact, self-contained)
\item \textit{``The U.S. Supreme Court has nine justices.''} (specific structural fact)
\end{itemize}

\paragraph{Bad Examples.}
\begin{itemize}
\item \textit{``Tesla's headquarters is in Austin, and Apple's headquarters is in Cupertino.''} (violates atomicity---two facts)
\item \textit{``The study shows that healthcare costs are rising.''} (violates unambiguity---which study?)
\item \textit{``This content appears on page 47 of the 10-K.''} (violates validity---trivial metadata)
\end{itemize}

\subsection{Valid Information Filtering}
\label{app:prompt-valid-filter}

\textbf{Prompt Design.} The prompt detects information completeness errors in atomic units. It filters out units that lack essential components, contain incomplete facts, or provide insufficient information for RAG applications. The prompt identifies five critical error patterns: (1) Missing essential elements (WHO/WHAT/WHEN/WHERE/HOW), (2) Fragmented information, (3) Contextually insufficient information, (4) Utility-deficient information (trivial facts, circular definitions), and (5) Meta information (document metadata, URLs, structural information). The output is a binary pass/fail decision with reasoning.

\paragraph{Good Examples (Pass).}
\begin{itemize}
\item \textit{``Albert Einstein discovered the theory of relativity in 1905.''} (complete: who, what, when)
\item \textit{``Disney announced its acquisition of 21st Century Fox for \$52.4 billion in December 2017.''} (all essential details)
\item \textit{``Amazon's quarterly sales increased by 13\% to \$143.1 billion in Q3 2023 compared to Q3 2022.''} (specific quantitative context)
\end{itemize}

\paragraph{Bad Examples (Fail).}
\begin{itemize}
\item \textit{``Discovered the theory of relativity.''} (missing who, when)
\item \textit{``The merger was announced.''} (missing which companies, when, financial terms)
\item \textit{``The document title is tesla\_2023\_annual\_report.pdf.''} (meta information with no substantive value)
\end{itemize}

\subsection{Valid Information Multi-Criteria Ranking}
\label{app:prompt-valid-rank}

\textbf{Prompt Design.} Five separate prompts rank atomic units across quality dimensions, each evaluated independently before combining with CRRF. The prompts instruct the LLM to provide numerical scores (0-1) with step-by-step reasoning.

\subsubsection{Criterion 1: Validity}

\textbf{Purpose.} Evaluates retrieval usefulness for RAG systems---whether information meaningfully supports answering user queries. Filters out overly abstract content, trivial notes, structural markup, or metadata-only fragments.

\paragraph{Good Examples.}
\begin{itemize}
\item \textit{``The Great Wall of China is over 21,000 kilometers long.''}
\item \textit{``Water boils at 100 degrees Celsius at standard atmospheric pressure.''}
\item \textit{``Shakespeare wrote the play Hamlet around 1600.''}
\end{itemize}

\paragraph{Bad Examples.}
\begin{itemize}
\item \textit{``Books contain pages with text.''} (trivial common knowledge)
\item \textit{``The sky appears during daytime.''} (obvious, low value)
\item \textit{``People use phones to make calls.''} (not meaningfully useful)
\end{itemize}

\subsubsection{Criterion 2: Completeness}

\textbf{Purpose.} Assesses whether information is self-contained and understandable without additional document context. Someone unfamiliar with the source should comprehend it independently.

\paragraph{Good Examples.}
\begin{itemize}
\item \textit{``The Amazon River is the second longest river in the world.''}
\item \textit{``Photosynthesis is the process by which plants convert sunlight into chemical energy.''}
\item \textit{``The capital city of Japan is Tokyo.''}
\end{itemize}

\paragraph{Bad Examples.}
\begin{itemize}
\item \textit{``It is the second longest river in the world.''} (missing subject)
\item \textit{``This process occurs primarily in the chloroplasts.''} (vague reference)
\item \textit{``The capital city there is Tokyo.''} (ambiguous location)
\end{itemize}

\subsubsection{Criterion 3: Specificity}

\textbf{Purpose.} Evaluates degree of concrete detail---presence of specific numbers, dates, names, entities, procedures, or conditions rather than vague generalities.

\paragraph{Good Examples.}
\begin{itemize}
\item \textit{``Mount Everest has a height of about 8,849 meters.''}
\item \textit{``The Constitution of the United States was ratified on September 17, 1787.''}
\item \textit{``DNA is composed of four bases: adenine, thymine, guanine, and cytosine.''}
\end{itemize}

\paragraph{Bad Examples.}
\begin{itemize}
\item \textit{``Mount Everest is very tall.''} (no specific measurement)
\item \textit{``The U.S. Constitution was created a long time ago.''} (vague temporal reference)
\item \textit{``DNA is made up of several components.''} (lacks enumeration)
\end{itemize}

\subsubsection{Criterion 4: Clarity}

\textbf{Purpose.} Ensures information is unambiguous with clear, unidirectional claims. Avoids polysemy, vague pronouns, or statements whose impact remains open to multiple interpretations.

\paragraph{Good Examples.}
\begin{itemize}
\item \textit{``Intense exercise increases the risk of accidents.''}
\item \textit{``A vegetarian diet reduces the risk of cardiovascular disease.''}
\item \textit{``Internet addiction lowers academic achievement.''}
\end{itemize}

\paragraph{Bad Examples.}
\begin{itemize}
\item \textit{``Intense exercise has a major impact.''} (unclear direction)
\item \textit{``A vegetarian diet brings many changes to health.''} (ambiguous effect)
\item \textit{``The internet has an important effect on society.''} (unspecified nature)
\end{itemize}

\subsubsection{Criterion 5: Questionability}

\textbf{Purpose.} Assesses whether information naturally forms the basis of a well-formed, specific question with a clear, factual, bounded answer suitable for retrieval-based QA.

\paragraph{Good Examples.}
\begin{itemize}
\item \textit{``Isaac Newton published \emph{Principia} in 1687.''} $\rightarrow$ When did Newton publish \emph{Principia}?
\item \textit{``Antarctica is the coldest continent on Earth.''} $\rightarrow$ Which is the coldest continent?
\item \textit{``The Berlin Wall collapsed in 1989.''} $\rightarrow$ When did the Berlin Wall collapse?
\end{itemize}

\paragraph{Bad Examples.}
\begin{itemize}
\item \textit{``\emph{Principia} was a profoundly important book.''} (too broad, no single answer)
\item \textit{``Antarctic exploration holds important significance in human history.''} (interpretive, open-ended)
\item \textit{``The collapse of the Berlin Wall was a major historical event.''} (not retrieval-friendly)
\end{itemize}

\subsection{Redundancy Detection}
\label{app:prompt-redundancy}

\textbf{Prompt Design.} The prompt determines whether multiple information units convey identical core facts regardless of surface-level differences in expression. It evaluates three types of equivalence: (1) \textbf{Semantic Equivalence}---same core meaning and conceptual content despite different wording, (2) \textbf{Logical Equivalence}---identical logical structure, reasoning patterns, and cause-effect relationships, and (3) \textbf{Factual Equivalence}---same entities, properties, relationships, and quantitative/qualitative aspects. The output is list of binary: REDUNDANT (information units are equivalent) or UNIQUE (distinct content).

\paragraph{REDUNDANT}
\begin{itemize}
\item \textit{Target: ``The company's revenue increased by 15\% in Q3 2023.''} \\ 
\textit{Comparison: ``In the third quarter of 2023, company revenues grew fifteen percent.''} \\
(Same core meaning, logical relationship, entity, time period, metric, and value)

\item \textit{Target: ``Water boils at 100 degrees Celsius at sea level.''} \\
\textit{Comparison: ``At standard atmospheric pressure, H$_2$O reaches boiling point at 100$^\circ$C.''} \\
(Equivalent despite different terminology)

\item \textit{Target: ``Shakespeare wrote Hamlet around 1600.''} \\
\textit{Comparison: ``The play Hamlet was authored by William Shakespeare circa 1600.''} \\
(Identical factual content, different expression)
\end{itemize}

\paragraph{UNIQUE}
\begin{itemize}
\item \textit{Target: ``The company's revenue increased by 15\% in Q3 2023.''} \\
\textit{Comparison: ``The company's expenses decreased by 10\% in Q3 2023.''} \\
(Different concepts: revenue vs. expenses; opposite directions: increase vs. decrease)

\item \textit{Target: ``Apple's iPhone sales grew in North America.''} \\
\textit{Comparison: ``Apple's iPad sales grew in North America.''} \\
(Different products despite same company and region)

\item \textit{Target: ``The patent was filed in 2023.''} \\
\textit{Comparison: ``The patent was granted in 2023.''} \\
(Different events: filing vs. granting)
\end{itemize}

\subsection{Question Generation}
\label{app:prompt-qgen}

\textbf{Prompt Design.} The prompt operates in two steps: (1) \textbf{Information Selection}---from a pool of atomic units, select items with highest connectivity potential for natural multi-hop reasoning, and (2) \textbf{Question Generation}---generate multiple unique questions requiring all selected units. The generation follows a two-level structure: \textbf{Level 1: Mandatory Logical Consistency} (zero-tolerance---all 4 criteria must pass) and \textbf{Level 2: Preferred Quality Dimensions} (for human preference ranking). Each generated question includes explicit reasoning checks for all 8 dimensions.

\subsubsection{Information Selection Criteria}

The prompt selects units based on: (1) Semantic connectivity---natural logical relationships or causal chains, (2) Complementary information---different pieces of a larger puzzle, (3) Reasoning chain potential---elegant paths vs. parallel facts, and (4) Natural integration---how humans naturally think about topics. Avoids completely independent facts, redundant information, or single-hop answerable items.

\subsubsection{Level 1: Mandatory Logical Consistency}

These four criteria mirror the Question Logical Filtering criteria (Section~\ref{app:prompt-qfilter}) and must all pass with zero tolerance:

\paragraph{1. Contextual Independence}
\begin{itemize}
\item \textit{Bad: ``According to the table above, which environmental policy is most effective?''}
\item \textit{Good: ``Which emissions reduction policy in the Inflation Reduction Act targets methane?''}
\end{itemize}

\paragraph{2. Answer Exclusion}
\begin{itemize}
\item \textit{Bad: ``Marie Curie worked at the University of Paris. When was the University of Paris established?''}
\item \textit{Good: ``What year was the university where Marie Curie worked established?''}
\end{itemize}

\paragraph{3. Information Equivalence}
\begin{itemize}
\item \textit{Bad (Overflow): ``What university did Marie Curie work at?''} (only needs one unit)
\item \textit{Bad (Underflow): ``How many students does the university where Marie Curie worked have?''} (requires external info)
\item \textit{Good: ``What year was the university where Marie Curie worked established?''} (requires exactly both units)
\end{itemize}

\paragraph{4. Question Clarity}
\begin{itemize}
\item \textit{Bad: ``Where is our headquarters located?''} (ambiguous pronoun)
\item \textit{Good: ``Where is Tesla's headquarters located?''} (clear entity reference)
\end{itemize}

\subsubsection{Level 2: Preferred Quality Dimensions}

These four dimensions mirror the Question Multi-Criteria Ranking criteria (Section~\ref{app:prompt-qrank}):

\paragraph{1. Connectivity}
\begin{itemize}
\item \textit{Bad: ``In what year was Apple founded, and in what year was the iPhone released?''} (parallel listing)
\item \textit{Good: ``How many years did it take Apple to release its first iPhone after its founding?''} (unified calculation)
\end{itemize}

\paragraph{2. Fluency}
\begin{itemize}
\item \textit{Bad: ``What is the temporal differential calculation when subtracting the institutional establishment year from 1900?''}
\item \textit{Good: ``When did Marie Curie receive her Nobel Prize in Chemistry?''}
\end{itemize}

\paragraph{3. Essentiality}
\begin{itemize}
\item \textit{Bad: ``What is the exact chronological period during which the highly distinguished Nobel laureate Marie Curie conducted her groundbreaking research at the prestigious University of Paris?''}
\item \textit{Good: ``When did Marie Curie work at the University of Paris?''}
\end{itemize}

\paragraph{4. Validity}
\begin{itemize}
\item \textit{Bad: ``What is the exact number of letters in the name of the university where Marie Curie worked?''}
\item \textit{Good: ``What element did Marie Curie discover and name after her native country?''}
\end{itemize}

\subsection{Question Logical Filtering}
\label{app:prompt-qfilter}

\textbf{Prompt Design.} Five filtering prompts implement zero-tolerance logical criteria. Each criterion must be satisfied independently; failure on any single criterion results in question rejection. The prompts enforce strict logical consistency to ensure questions are well-formed for RAG evaluation.

\subsubsection{Criterion 1: Contextual Independence}

\textbf{Purpose.} Detects questions that improperly assume or reference external document structures, sentence numbering, or meta-textual elements. Questions must be self-contained and understandable without knowledge of their presentation format.

\paragraph{Good Examples (Pass).}
\begin{itemize}
\item \textit{``What are the primary effects of climate change on coastal ecosystems?''}
\item \textit{``Which renewable energy sources have the highest efficiency ratings?''}
\item \textit{``What methods are most effective for carbon emission reduction?''}
\end{itemize}

\paragraph{Bad Examples (Fail).}
\begin{itemize}
\item \textit{``What climate change effects are mentioned in the second sentence?''} (sentence numbering reference)
\item \textit{``Based on the document provided, what are the primary renewable energy sources?''} (document reference)
\item \textit{``According to the table above, which environmental policy is most effective?''} (structural reference)
\end{itemize}

\subsubsection{Criterion 2: Answer Exclusion}

\textbf{Purpose.} Identifies questions that embed partial or complete answers within the question formulation, eliminating genuine multi-hop reasoning requirements. Questions must not contain pre-provided intermediate reasoning steps.

\paragraph{Good Examples (Pass).}
\begin{itemize}
\item \textit{``What year was the university where Marie Curie worked established?''} (requires two-hop: identify university, then find establishment date)
\item \textit{``What year was the founder of Microsoft born?''} (requires: identify founder, then find birth year)
\item \textit{``Which acquisition did Tesla's CEO complete in 2022?''} (requires: identify CEO, then find acquisition)
\end{itemize}

\paragraph{Bad Examples (Fail).}
\begin{itemize}
\item \textit{``Marie Curie worked at the University of Paris. When was the University of Paris established?''} (embeds first hop answer)
\item \textit{``The founder of Microsoft, Bill Gates, was born in which year?''} (reveals founder identity)
\item \textit{``Tesla's CEO Elon Musk made which acquisition in 2022?''} (provides CEO name)
\end{itemize}

\subsubsection{Criterion 3: Information Equivalence}

\textbf{Purpose.} Enforces the Exact Information Equivalence Equation: questions must be fully answerable using only the provided atomic units and must require all units without redundancy. Detects two error patterns: (1) \textbf{Information Overflow}---questions solvable with only a subset of provided units, and (2) \textbf{Information Underflow}---questions requiring external knowledge beyond provided units.

\paragraph{Good Examples (Pass).}
\begin{itemize}
\item \textit{Provided: ``Microsoft was founded by Bill Gates,'' ``Bill Gates was born in 1955,'' ``Microsoft headquarters is in Redmond, Washington''} \\
\textit{Question: ``Where is Microsoft's headquarters located, and when was the founder born?''} \\
(Requires all three units; no overflow or underflow)

\item \textit{Provided: ``Apple Inc. was founded in 1976,'' ``Steve Jobs co-founded Apple with Steve Wozniak,'' ``Apple's HQ is in Cupertino''} \\
\textit{Question: ``When was the company co-founded by Steve Jobs and Steve Wozniak established, and where is it headquartered?''} \\
(Information Required = Information Provided exactly)
\end{itemize}

\paragraph{Bad Examples (Fail).}
\begin{itemize}
\item \textit{Provided: ``Microsoft was founded by Bill Gates,'' ``Bill Gates was born in 1955,'' ``Microsoft HQ is in Redmond''} \\
\textit{Question: ``Who founded Microsoft?''} \\
(Information Overflow---only first unit needed; units 2 and 3 redundant)

\item \textit{Provided: ``Tesla released its first EV in 2008,'' ``Elon Musk is CEO of Tesla''} \\
\textit{Question: ``What was Tesla's stock price when it went public?''} \\
(Information Underflow---IPO information not provided)
\end{itemize}

\subsubsection{Criterion 4: Question Clarity}

\textbf{Purpose.} Detects questions with vague referential terms, underspecified comparisons, or context-dependent queries. All referential terms must have clear antecedents, and comparisons must include explicit reference points.

\paragraph{Good Examples (Pass).}
\begin{itemize}
\item \textit{``Who directed the 1994 film `The Lion King'?''} (specific movie reference)
\item \textit{``Between Steven Spielberg and Martin Scorsese, who has directed more feature films to date?''} (explicit comparison basis)
\item \textit{``What factors influenced Apple's decision to discontinue the iPod Classic in 2014?''} (clear entity and event)
\end{itemize}

\paragraph{Bad Examples (Fail).}
\begin{itemize}
\item \textit{``That movie was directed by whom?''} (vague reference---which movie?)
\item \textit{``Who worked in the field longer?''} (underspecified---which field? compared to whom?)
\item \textit{``When was the company more profitable?''} (ambiguous---which company? compared to when?)
\end{itemize}

\subsubsection{Criterion 5: Answerability}

\textbf{Purpose.} Determines whether questions can be completely answered using only the provided document chunks. Questions must have all necessary information present without requiring external knowledge or critical information gaps.

\paragraph{Good Examples (Pass).}
\begin{itemize}
\item \textit{Chunks contain: revenue figures, growth percentages, time periods} \\
\textit{Question: ``What was the company's revenue growth rate in Q3 2023?''} \\
(All information present for complete answer)

\item \textit{Chunks contain: founding date, founder names, company location} \\
\textit{Question: ``When and where was the company founded, and by whom?''} \\
(Complete information available across chunks)
\end{itemize}

\paragraph{Bad Examples (Fail).}
\begin{itemize}
\item \textit{Chunks contain: product launch date, product features} \\
\textit{Question: ``What was the market share after the product launch?''} \\
(Market share data missing from chunks)

\item \textit{Chunks contain: company name, CEO name} \\
\textit{Question: ``What was the CEO's previous work experience before joining?''} \\
(Previous employment history not in chunks)
\end{itemize}

\subsection{Question Multi-Criteria Ranking}
\label{app:prompt-qrank}

\textbf{Prompt Design.} Four ranking prompts evaluate candidate questions across quality dimensions. Each prompt provides numerical scores (0-1) with reasoning. Rankings are combined using CRRF to select the final question.

\subsubsection{Criterion 1: Connectivity}

\textbf{Purpose.} Measures logical flow and integration elegance in multi-hop reasoning. Evaluates how naturally the question connects multiple information sources through four dimensions: (1) Logical flow smoothness---natural transitions between reasoning steps, (2) Answer synthesis quality---coherent unified result vs. parallel enumeration, (3) Reasoning chain elegance---intellectually satisfying path, and (4) Information integration depth---meaningful semantic connections.

\paragraph{Good Examples (High Connectivity).}
\begin{itemize}
\item \textit{``By how much did Microsoft's cloud revenue grow from the year Satya Nadella became CEO to FY2024?''} (smooth temporal reasoning with unified result)
\item \textit{``How many years did it take Apple to release its first iPhone after its founding?''} (elegant temporal calculation)
\item \textit{``How many times larger is China's population compared to the U.S. population?''} (meaningful quantitative integration)
\end{itemize}

\paragraph{Bad Examples (Low Connectivity).}
\begin{itemize}
\item \textit{``What were Microsoft's revenues in FY2023 and FY2024, respectively?''} (parallel enumeration, no synthesis)
\item \textit{``In what year was Apple founded, and in what year was the iPhone released?''} (choppy listing)
\item \textit{``What is the population of China, and when was the U.S. Constitution ratified?''} (superficial connection)
\end{itemize}

\subsubsection{Criterion 2: Fluency}

\textbf{Purpose.} Measures natural readability and conversational quality. Evaluates whether the question sounds like something a real person would naturally ask through two dimensions: (1) Naturalness---conversational authenticity without artificial template-like phrasing, and (2) Expression appropriateness---accessible vocabulary and straightforward syntax with creative rephrasing.

\paragraph{Good Examples (High Fluency).}
\begin{itemize}
\item \textit{``When did Marie Curie receive her Nobel Prize in Chemistry?''}
\item \textit{``In which city is Amazon---the e-commerce company founded by Jeff Bezos---headquartered?''}
\item \textit{``How much did The Walt Disney Company pay to acquire the media group founded by Rupert Murdoch?''}
\end{itemize}

\paragraph{Bad Examples (Low Fluency).}
\begin{itemize}
\item \textit{``What is the temporal differential calculation when subtracting the institutional establishment year from the chronological marker of 1900?''} (unnecessarily complex)
\item \textit{``Identify the precise geographical placement of Amazon's principal global operational headquarters.''} (artificial phrasing)
\item \textit{``State the capital monetary figure associated with Disney's large-scale corporate acquisition.''} (overly formal)
\end{itemize}

\subsubsection{Criterion 3: Essentiality}

\textbf{Purpose.} Measures focus on core information while eliminating unnecessary elements. Evaluates modifier restraint---absence of excessive adjectives, adverbs, and decorative expressions that LLMs often generate.

\paragraph{Good Examples (High Essentiality).}
\begin{itemize}
\item \textit{``When did Marie Curie work at the University of Paris?''} (clean and factual)
\item \textit{``How much did Disney pay to acquire 21st Century Fox?''} (no unnecessary modifiers)
\item \textit{``What are the four bases that compose DNA?''} (direct and concise)
\end{itemize}

\paragraph{Bad Examples (Low Essentiality).}
\begin{itemize}
\item \textit{``What is the exact chronological period during which the highly distinguished Nobel laureate Marie Curie conducted her groundbreaking research at the prestigious University of Paris?''} (excessive modifiers)
\item \textit{``What was the exact amount of the historically significant deal when Disney announced its acquisition of the giant global media company 21st Century Fox?''} (decorative language)
\item \textit{``What are the four crucial fundamental elements that make up DNA, the intricate and sophisticated blueprint of genetics?''} (redundant descriptors)
\end{itemize}

\subsubsection{Criterion 4: Validity}

\textbf{Purpose.} Measures substantive worth and rational motivation. Evaluates through two dimensions: (1) Information worth---genuine insight beyond trivial fact-checking with educational/intellectual value, and (2) Query motivation---clear rational reason arising from natural curiosity or legitimate inquiry needs.

\paragraph{Good Examples (High Validity).}
\begin{itemize}
\item \textit{``What is considered Marie Curie's greatest achievement?''} (meaningful historical insight)
\item \textit{``Which scientists discovered the structure of DNA in 1953?''} (educational value)
\item \textit{``What is the theoretical maximum efficiency of a silicon single-junction solar cell?''} (practical knowledge)
\end{itemize}

\paragraph{Bad Examples (Low Validity).}
\begin{itemize}
\item \textit{``What is the exact number of letters in the name of the university where Marie Curie worked?''} (trivial numerical fact)
\item \textit{``What is the title of this document?''} (no substantive insight)
\item \textit{``On which page can the comprehensive report be found?''} (meta information, no value)
\end{itemize}

\end{document}